\PassOptionsToPackage{table,dvipsnames}{xcolor}
\documentclass{pediamed-ai}

\title{Toward Cognitive Supersensing in Multimodal Large Language Models}

\author[1,*]{Boyi Li}
\author[1,*,\S]{Yifan Shen}
\author[1,*]{Yuanzhe Liu}
\author[1]{Yifan Xu}
\author[1]{Jiateng Liu}
\author[1]{Xinzhuo Li}
\author[1]{Zhengyuan Li}
\author[2]{Jingyuan Zhu}
\author[1]{Yunhan Zhong}
\author[2]{Fangzhou Lan}
\author[2]{Jianguo Cao}
\author[1]{James M. Rehg}
\author[1]{Heng Ji}
\author[1,\dagger]{Ismini Lourentzou}
\author[1,2,\dagger]{Xu Cao}

\affiliation[1]{University of Illinois Urbana-Champaign}
\affiliation[2]{PediaMed AI}

\contribution[*]{Equal contribution}
\contribution[\S]{Project lead}
\contribution[\dagger]{Corresponding author}

\usepackage[utf8]{inputenc} %
\usepackage{url}            %
\usepackage{amsfonts}       %
\usepackage{nicefrac}       %

\usepackage{mathtools}
\usepackage{amsthm}

\usepackage{tabularx}
\usepackage{ltablex}
\usepackage{tabularray}
\newcolumntype{C}{>{\centering\arraybackslash}X}

\usepackage{diagbox}
\usepackage{hhline}
\usepackage{makecell}
\usepackage{array}
\usepackage{adjustbox}
\usepackage{longtable}

\usepackage{color}

\usepackage{colortbl}

\usepackage{epsfig}
\usepackage{wrapfig}
\usepackage{needspace}
\usepackage{floatflt}

\usepackage{enumerate}
\usepackage{enumitem}
\usepackage{pifont}
\usepackage{xspace}
\usepackage{multicol}
\usepackage{changepage}

\usepackage{soul}
\usepackage{siunitx}
\usepackage{calc}

\usepackage{wasysym}

\usepackage{algorithm}
\usepackage{algorithmic}

\usepackage{scalefnt}

\crefname{appendix}{Appendix}{Appendices}
\crefname{section}{Section}{Sections}
\crefname{figure}{Figure}{Figures}
\crefname{table}{Table}{Tables}

\usepackage{titlecaps}

\newcolumntype{L}[1]{>{\raggedright\let\newline\\\arraybackslash\hspace{0pt}}m{#1}}
\newcolumntype{R}[1]{>{\raggedleft\let\newline\\\arraybackslash\hspace{0pt}}m{#1}}

\newcommand{\ignore}[1]{}

\makeatletter
\DeclareRobustCommand\onedot{\futurelet\@let@token\@onedot}
\def\@onedot{\ifx\@let@token.\else.\null\fi\xspace}

\def\ie{i.e\onedot}

\makeatother

\definecolor{MyBlue}{rgb}{0.46, 0.50, 0.61}
\definecolor{MyDarkBlue}{rgb}{0,0.08,0.8}
\definecolor{MyDarkGreen}{RGB}{45,155,45}
\definecolor{MyDarkRed}{rgb}{0.8,0.02,0.02}
\definecolor{MyOrange}{rgb}{1.0, 0.4, 0.2}
\definecolor{MyPurple}{RGB}{111,0,255}
\definecolor{MyRed}{rgb}{0.8,0.0,0.0}
\definecolor{MyGold}{rgb}{0.75,0.6,0.12}
\definecolor{MyDarkgray}{rgb}{0.66, 0.66, 0.66}
\definecolor{MyBrown}{rgb}{0.65, 0.16, 0.16}
\definecolor{MyMutedRose}{rgb}{0.58, 0.29, 0.35}
\definecolor{JiayuanColor}{rgb}{0.60,0.43,0.48}
\definecolor{erranColor}{rgb}{24, 40, 113}

\definecolor{citecolor}{HTML}{696FAD}

\definecolor{bggray}{HTML}{F5F5F5}
\definecolor{pvdblue}{HTML}{DAE8FC}
\definecolor{RoseQuartzBg}{HTML}{F7CAC9}
\definecolor{RoseQuartz}{HTML}{F5A798}
\definecolor{Serenity}{HTML}{92A8D1}
\definecolor{OrangeRed}{rgb}{1.0, 0.27, 0.0}
\definecolor{RoyalBlue}{cmyk}{1, 0.50, 0, 0}
\definecolor{Turquoise}{HTML}{0F4C81}
\definecolor{mint}{rgb}{0.24, 0.71, 0.54}
\definecolor{green}{rgb}{0.0, 0.120, 0.0}

\newdimen\abovecrulesep
\newdimen\belowcrulesep
\makeatletter
\patchcmd{\@@@cmidrule}{\aboverulesep}{\abovecrulesep}{}{}
\patchcmd{\@xcmidrule}{\belowrulesep}{\belowcrulesep}{}{}
\makeatother

\definecolor{mybluetitle}{HTML}{4B527E} %

\definecolor{codegreen}{HTML}{478058}%
\definecolor{codegray}{rgb}{0.5,0.5,0.5}
\definecolor{codepurple}{HTML}{4F5E80} %
\definecolor{backcolour}{rgb}{0.95,0.95,0.92}
\lstdefinestyle{mystyle}{
    backgroundcolor=\color{backcolour},
    commentstyle=\color{codegreen},
    keywordstyle=\color{magenta},
    numberstyle=\tiny\color{codegray},
    stringstyle=\color{codepurple},
    basicstyle=\ttfamily\scriptsize,
    breakatwhitespace=false,
    breaklines=true,
    captionpos=b,
    keepspaces=true,
    frame=none,
    numbersep=5pt,
    showspaces=false,
    showstringspaces=false,
    showtabs=false,
    tabsize=2
}

\newtcolorbox{promptbox}[2][]{
    enhanced,
    breakable,
    center title,
    left*=0pt, right*=0pt,
    boxsep=2pt, left=5pt, right=5pt,
    skin first=enhanced,
    skin middle=enhanced,
    skin last=enhanced,
    colback  = backcolour,
    fonttitle=\bfseries\rmfamily,
    fontupper=\scriptsize,
    title={\footnotesize\strut{#2}},
    #1
    }

\newtcolorbox{onebox}[2][]{
    enhanced,
    center title,
    left*=0pt, right*=0pt,
    boxsep=2pt, left=5pt, right=5pt,
    skin first=enhanced,
    skin middle=enhanced,
    skin last=enhanced,
    colframe = mybluetitle!90,
  colback  = mybluetitle!10,
    fonttitle=\bfseries\rmfamily\fontfamily{phv}\selectfont,
    title={\footnotesize\strut{#2}  \refstepcounter{subsubsection} \addcontentsline{toc}{subsubsection}{\string\numberline{\thesubsubsection}#2}
    },
    #1
    }

\usepackage{amsmath,amsfonts,bm}

\def\eqref#1{equation~\ref{#1}}

\def\1{\bm{1}}

\DeclareMathAlphabet{\mathsfit}{\encodingdefault}{\sfdefault}{m}{sl}
\SetMathAlphabet{\mathsfit}{bold}{\encodingdefault}{\sfdefault}{bx}{n}

\newcommand{\ourmodel}{{CogSense-8B}\xspace}
\newcommand{\ourmethod}{{Cognitive Supersensing}\xspace}
\newcommand{\Enc}{\mathrm{Enc}}
\newcommand{\vis}{\mathrm{vis}}
\newcommand{\txt}{\mathrm{txt}}
\newcommand{\opt}{\mathrm{opt}}
\newcommand{\SFT}{\mathrm{SFT}}
\newcommand{\MSE}{\operatorname{MSE}}

\newcommand{\eg}{\textit{e.g.},\xspace}      %
\renewcommand{\paragraph}[1]{%
	\noindent\textbf{#1.}\noindent\xspace%
}

\newcommand{\kindatiny}{\fontsize{6pt}{7.2pt}\selectfont}
\newlength\savewidth

\definecolor{qcolor}{HTML}{536872}

\newcolumntype{P}[1]{>{\centering\arraybackslash}p{#1}}

\newcommand{\tablestyle}[2]{%
	\fontfamily{ptm}\selectfont%
	\let\itold\it%
	\def\it{\itold \fontfamily{ptm}\selectfont}%
	\setlength{\tabcolsep}{#1}\renewcommand{\arraystretch}{#2}\centering\kindatiny%
	\let\citeold\cite%
	\renewcommand{\cite}[1]{\normalfont\fontfamily{ptm}\selectfont\tiny\citeold{##1}}%
}
\newcolumntype{x}[1]{>{\centering\arraybackslash}p{#1pt}}
\newcolumntype{y}[1]{>{\raggedright\arraybackslash}p{#1pt}}
\newcolumntype{z}[1]{>{\raggedleft\arraybackslash}p{#1pt}}
\newcolumntype{u}{>{\centering\arraybackslash}p{20pt}}
\newcolumntype{a}{>{\centering\arraybackslash}p{16pt}}
\newcolumntype{Y}{>{\centering\arraybackslash}X}

\definecolor{logoRed}{HTML}{CB2F10}
\definecolor{logoBlue}{HTML}{1A4E8A}
\definecolor{logoCyan}{HTML}{56BBCC}

\definecolor{genLLM}{RGB}{250,250,250}
\definecolor{specLLM}{RGB}{230,230,230}
\definecolor{ourLLM}{RGB}{230,242,230}
\definecolor{spaLLM}{RGB}{255,249,196}

\definecolor{genVLM}{RGB}{255,255,255}
\definecolor{ourVLM}{RGB}{210,210,210}

\definecolor{lightgray}{rgb}{0.95, 0.95, 0.95}
\definecolor{baselinecolor}{gray}{.9}

\tcbset{colback=white}
\colorlet{titleblue}{blue!80!black}
\colorlet{titlered}{red!80!black}
\colorlet{titlegreen}{green!80!black}
\colorlet{darkgreen}{green!50!black}

\newtcolorbox{planbox}[1]{
    colback=gray!5,
    colframe=gray!75,
    title=#1,
    fonttitle=\bfseries
}

\tcbset{
  aibox/.style={
    width=\linewidth,
    top=10pt,
    colback=white,
    colframe=black,
    colbacktitle=black,
    enhanced,
    center,
    attach boxed title to top left={yshift=-0.1in,xshift=0.15in},
    boxed title style={boxrule=0pt,colframe=white,},
  }
}

\newtcolorbox{AIbox}[2][]{aibox,title=#2,#1}

\definecolor{applegreen}{rgb}{0.55, 0.71, 0.0}

\newcommand{\red}[1]{$_{\color{RedOrange}\uparrow #1}$}

\abstract{Multimodal Large Language Models (MLLMs) have made rapid progress in open-vocabulary visual understanding, yet they often struggle with cognitive visual reasoning tasks that require maintaining, transforming, and simulating abstract visual states. Existing approaches primarily scale Chain-of-Thought (CoT) reasoning in language, forcing inherently spatial or visual computations into discrete text tokens. We introduce {Cognitive Supersensing}, a training paradigm that equips MLLMs
with latent visual reasoning by integrating a Latent Visual Imagery Prediction (LVIP) head that progressively learns refined visual latent states aligned with the evolving reasoning context and target answer, thereby
providing a structured visual grounding signal for internal reasoning. We further introduce a reinforcement learning stage that optimizes rationale trajectories using both answer evidence and LVIP-based latent grounding, encouraging reasoning paths that remain visually consistent. To evaluate the cognitive capabilities of MLLMs, we present CogSense-Bench, a comprehensive visual question answering (VQA) benchmark assessing five cognitive dimensions. Extensive experiments demonstrate that MLLMs trained with Cognitive Supersensing significantly outperform state-of-the-art baselines on CogSense-Bench and exhibit superior generalization on out-of-domain mathematics and science VQA benchmarks, suggesting that internal visual imagery may be key to bridging the gap between perceptual recognition and cognitive understanding. Causal analyses through noise injection, latent patching, and linear probing show that LVIP states progressively encode task-relevant visual structure and influence downstream answer probabilities, supporting their role as active reasoning signals.}

\metadata[Project page]{\url{https://pediamed.ai/Cognition-MLLM/cogsense/}}
\metadata[GitHub]{\url{https://github.com/PediaMedAI/Cognition-MLLM}}
\metadata[HuggingFace]{\url{https://huggingface.co/datasets/PediaMedAI/CogSense-Bench}}
\metadata[Correspondence to]{\email{lourent2@illinois.edu}, \email{xucao@pediamed.ai}}

\definecolor{citecolor}{HTML}{0071BC}
\hypersetup{
  citecolor=citecolor,
  pdftitle={Toward Cognitive Supersensing in Multimodal Large Language Models},
  pdfauthor={Boyi Li, Yifan Shen, Yuanzhe Liu, Yifan Xu, Jiateng Liu, Xinzhuo Li, Zhengyuan Li, Jingyuan Zhu, Yunhan Zhong, Fangzhou Lan, Jianguo Cao, James M. Rehg, Heng Ji, Ismini Lourentzou, Xu Cao}
}

\begin{document}
\maketitle

\section{Introduction}
\label{sec:intro}

Multimodal Large Language Models (MLLMs) have rapidly advanced open-vocabulary visual understanding, enabling strong performance across recognition, grounding, and compositional perception tasks~\citep{yang2025thinking}. Yet, existing models can describe what is present but struggle to mentally operate on visual information, \eg by explaining abstract layouts, simulating future transformations, or inferring visual rules in diagrams~\citep{schulze2025visual}. These failures expose a gap between high-level cognitive reasoning and low-level perception; this gap has become a key bottleneck in modern VQA and visual reasoning benchmarks~\citep{lu2024mathvista,cao2025visualcognitiongaphumans} and even extends to reasoning-driven image generation~\citep{liu2026thinking,susladkar2026best}.\looseness-1

Current approaches often attempt to narrow this gap by eliciting \emph{Chain-of-Thought} (CoT) reasoning as natural-language reasoning traces, encouraging models to articulate intermediate steps explicitly \citep{wei2023chainofthoughtpromptingelicitsreasoning,zhang2024multimodalchainofthoughtreasoninglanguage,tabassum2025mmplanner}, or by relying on external tools~\citep{hu2024visual}. However, even in multimodal settings, the intermediate computation is externalized in text, which is a poor interface for many visuospatial operations. Many subroutines underlying abstract visual reasoning, such as mentally rotating shapes, simulating dynamics, or inducing rules in pattern matrices, are most naturally expressed as geometric transformations, continuous states, or structured visual relations rather than as a sequence of discrete tokens. Expressing these intermediate states solely in linear text can introduce representational bottlenecks, where spatial relations are compressed into discrete tokens, increasing the risk of information loss and brittle reasoning.\looseness-1

To quantify this cognitive gap, we introduce \textbf{CogSense-Bench}, a comprehensive benchmark that operationalizes visual cognition along five core dimensions: fluid intelligence~\citep{cattell1963theory, gentner1983structure}, crystallized intelligence~\citep{cattell1963theory, rosch1973natural}, visuospatial cognition~\citep{wertheimer1923untersuchungen, Biederman1987Recognition}, mental simulation~\citep{Battaglia2013simulation}, and visual routines~\citep{ullman1984visual}, which together underpin intuitive theories of physics and psychology~\citep{lake2017building}. Using CogSense-Bench, we systematically re-evaluate state-of-the-art (SoTA) MLLMs and find that substantial weaknesses persist across all dimensions, even when models are equipped with CoT prompting, suggesting that text-only reasoning is indeed often a brittle interface for tasks that require multi-step visual transformation and manipulation.

We therefore seek to shift part of the intermediate reasoning from discrete tokens to a representation space that better preserves geometry, continuity, and structured visual relations. This direction is consistent with cognitive science accounts of a visuospatial sketchpad (the ``mind's eye'') that supports maintaining and transforming internal visual representations during problem solving~\citep{ganis2011visual,xie2020visual,tabi2022vividness}, and with recent progress showing that latent visual states can serve as effective substrates for prediction and world modeling~\citep{yang2025cambriansspatialsupersensingvideo,shen2026phantom}, as well as with brain-inspired designs that draw on hippocampal mechanisms to enhance video vision language models~\citep{cao2026enhancing}. Motivated by these insights, we explore whether equipping MLLMs with an internal visual reasoning substrate can better support multi-step, non-linguistic manipulation of visual information. To this end, we introduce \textbf{Cognitive Supersensing}, a training paradigm that encourages MLLMs to construct latent image-based internal reasoning chains. Concretely, we augment an MLLM with a \textbf{Latent Visual Imagery Prediction (LVIP)} head that predicts progressively refined latent visual representations, \ie latent ``imagery'' states that encode intermediate visual reasoning, and aligns them with image representations derived from the target answer and supervision signal, thereby grounding the latent chain toward the correct solution. This design enables multi-step reasoning in a representation space that better preserves visual structure, while still allowing language to provide high-level semantic guidance and final explanations.\looseness-1

To train models under this paradigm, we further curate the train split of CogSense-Bench, a targeted training set spanning five categories and eleven sub-tasks, and adopt a multi-stage training pipeline. First, in a supervised fine-tuning stage, we optimize an \textbf{LVIP} objective that provides an auxiliary learning signal encouraging intermediate latent visual states to stay predictive of the final answer's visual representation, tightly coupling the reasoning process with the solution. Second, we introduce a reinforcement learning stage that optimizes rollout trajectories in a latent space conditioned on these visual cognitive latents, encouraging coherent latent dynamics and discouraging brittle text-only reasoning paths. Across extensive experiments, \ourmodel achieves SoTA performance on all tasks and the strongest overall accuracy (\textbf{73.8}), surpassing GPT-5.2 by \textbf{+33.5}, while also demonstrating improvements in out-of-domain (OOD) generalization, outperforming strong baselines on challenging science and open-ended VQA benchmarks.
Taken together, our results suggest that latent visual imagery can serve as an effective internal scaffold that bridges perceptual recognition and cognitive-level multimodal reasoning.
Overall, our contributions are:
\begin{itemize}[itemsep=0.5ex, parsep=0pt, topsep=-2.3pt, leftmargin=0.4cm]
    \item We propose \textbf{Cognitive Supersensing}, a latent-space reasoning-and-learning framework with supervised fine-tuning and reinforcement learning stages that endows MLLMs with visual imagery capability by aligning semantic reasoning with latent visual world modeling.
    \item We introduce \textbf{CogSense-Bench}, a comprehensive benchmark spanning five cognitive dimensions: fluid intelligence, crystallized intelligence, visuospatial cognition, mental simulation, and visual routines, providing a systematic testbed for evaluating visual cognition beyond perceptual recognition and supporting future research in this direction.
    \item Through extensive experiments, we show that reasoning and planning purely in text space is often insufficient for visual cognition: \ourmodel achieves SoTA performance on CogSense-Bench and exhibits strong OOD generalization on challenging mathematics and science VQA benchmarks, outperforming competitive baselines that rely on text-only planning.
\end{itemize}

\section{Related Work}
\label{sec:related}

Our work is closely related to Abstract Reasoning~\citep{chollet2019measure, legris2024harcrobustestimatehuman, chollet2025arcagi2newchallengefrontier}, Social Reasoning~\citep{lee2024towards,cao2026gaze,cao2026toward,cao2025socialgesture}, Visual Cognition~\citep{bongard1968recognition, bongard1970pattern, raven1936mental, raven1998raven, raven2003raven, barrett2018measuring, zhang2019raven}, VLMs for Visual Reasoning~\citep{radford2021learningtransferablevisualmodels, gupta2022visualprogrammingcompositionalvisual, li2023blip2bootstrappinglanguageimagepretraining, bai2025qwen25vltechnicalreport, bai2025qwen3vltechnicalreport, liu2024improved, wang2024cogvlmvisualexpertpretrained}, and Latent Visual Reasoning~\citep{yang2025machinementalimageryempower, zhang2025latentsketchpadsketchingvisual, li2025latentvisualreasoning, li2025latentimplicitvisualreasoning, wang2024reconstructive}. Cognitive Supersensing differs from prior reconstructive and latent-visual-reasoning methods by making latent visual states part of the reasoning trajectory. Instead of reconstructing the input image or using visual latents as generic intermediate scratchpads, LVIP predicts visual hypotheses conditioned on intermediate reasoning prefixes, and the latent-rationale optimization stage uses their consistency to shape the reasoning policy. Additional related work is discussed in Appendix~\ref{appx_sec:related}. \looseness=-1

\begin{figure*}[!p]
    \centering
    \includegraphics[width=0.99\linewidth]{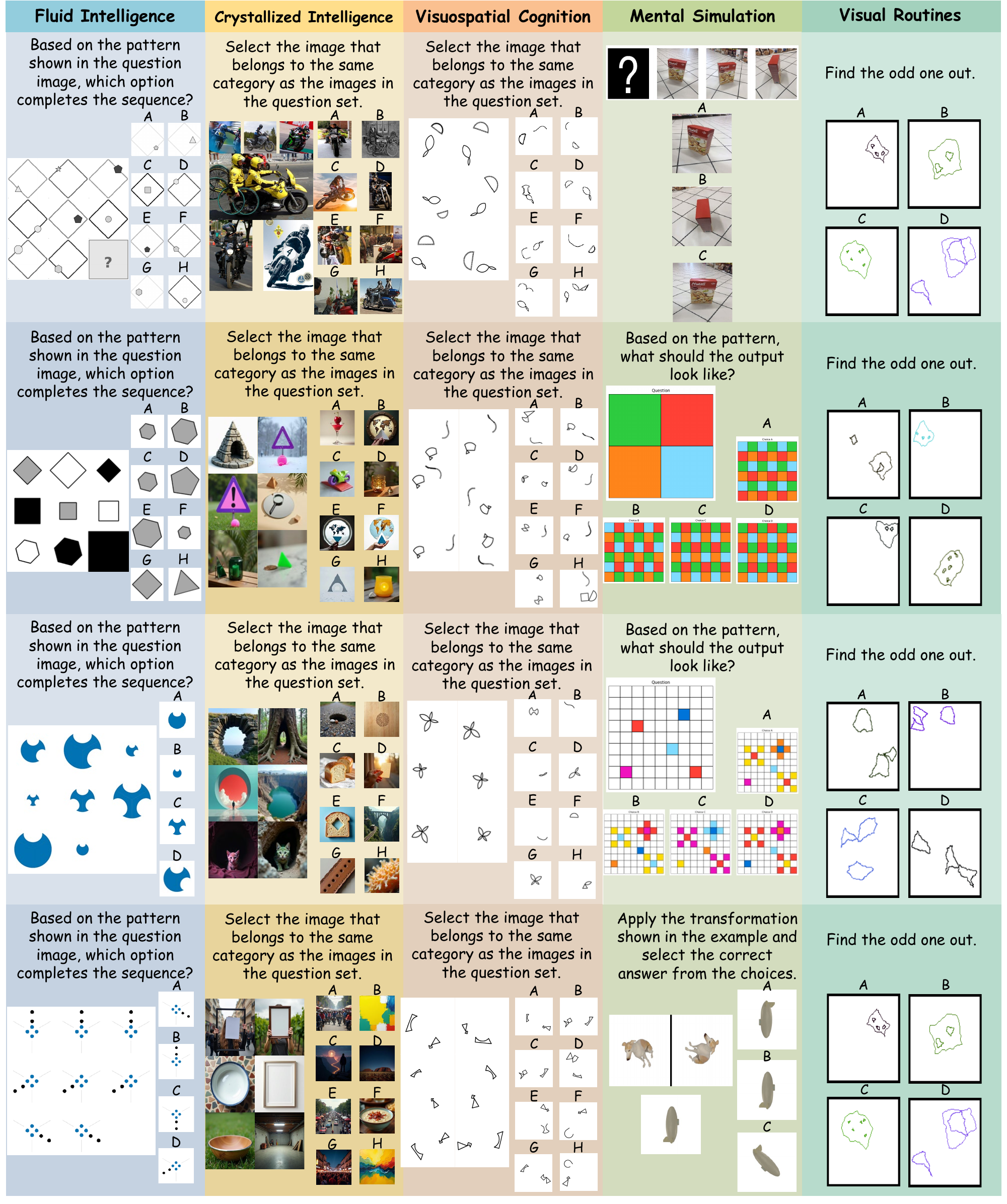}
   
    \caption{\textbf{CogSense-Bench Examples.} Samples from each category of CogSense-Bench. CogSense-Bench comprises various visual cognitive questions classified into five categories: Fluid Intelligence, Crystallized Intelligence, Visuospatial Cognition, Mental Simulation, and Visual Routines, which require visual imagery and cognitive supersensing with deep thinking and reasoning.}
    \label{fig:dataset_example}
\end{figure*}

\section{CogSense Benchmark}
\label{sec:dataset}

\begin{figure*}[!t]
    \centering
    \includegraphics[width=0.99\linewidth]{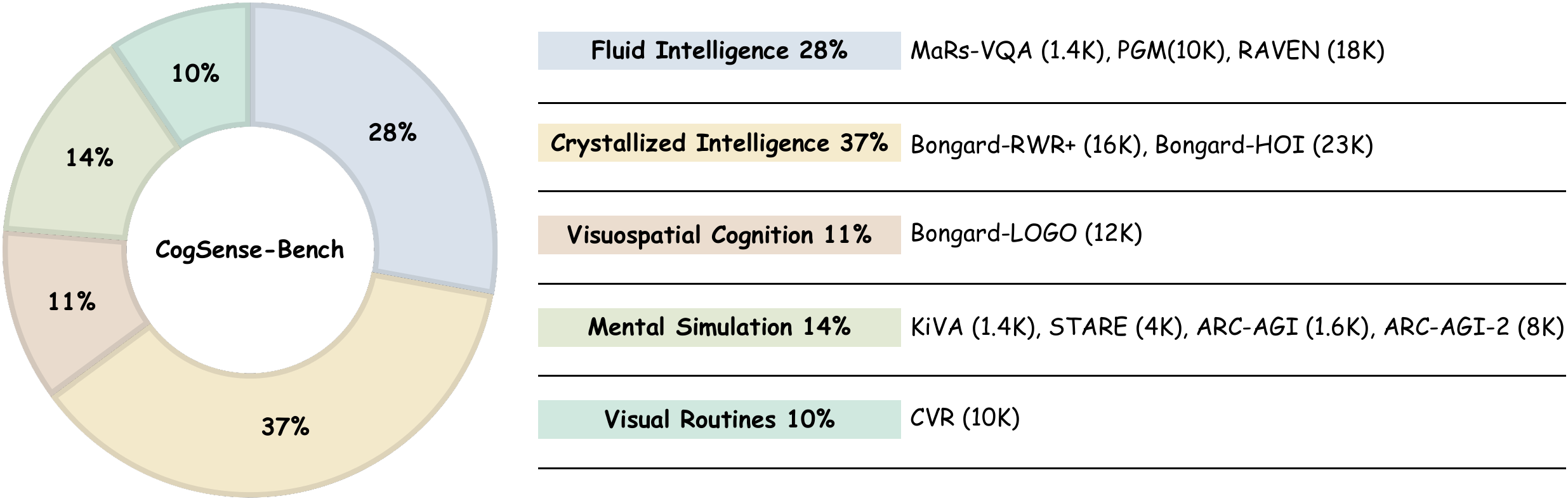}
     
    \caption{\textbf{Data Distribution of CogSense-Bench.} The dataset is structured around five human-cognitive capabilities rather than traditional vision tasks. By aggregating complex abstract reasoning benchmarks, it specifically targets higher-order visual logic and System-2 thinking.}
    \label{fig:dataset_satas}
     
\end{figure*}

While numerous VQA benchmarks evaluate MLLMs across various tasks, they predominantly focus on semantic recognition or description~\citep{li2025unfoldingspatialcognitionevaluating, chia2024puzzlevqadiagnosingmultimodalreasoning}. There remains a lack of systematic evaluation protocols specifically designed to assess high-level visual cognition~\citep{shen2026evaluating,shen2026position}, \ie the ability to reason over visual inputs through distinct cognitive mechanisms such as abstract reasoning, spatial structuring, mental simulation, and attention-driven operations. To address this gap, we introduce \textbf{CogSense-Bench}, which consists of a comprehensive and large-scale train split together with a unified multi-task test split. \looseness=-1

Visual cognition supports high-level visual reasoning and inference processes, such as causal reasoning, intuitive physics, and intuitive psychology~\citep{schulze2025visual, Battaglia2013simulation}. These capacities reflect central dimensions through which humans interpret, predict, and reason about visual environments. To systematically evaluate whether MLLMs exhibit analogous competencies, we assess visual cognitive intelligence across five theory-grounded cognitive categories. These categories are selected to align with established constructs in cognitive science and psychology, and to capture complementary mechanisms underlying human visual reasoning as articulated in prior theoretical and empirical work.

\textbf{Fluid Intelligence} evaluates the capacity to solve unseen reasoning problems independently of prior knowledge~\citep{cattell1963theory}. Grounded in Structure Mapping Theory~\citep{gentner1983structure}, this requires the model to transcend surface attributes and map higher-order logical rules.

\textbf{Crystallized Intelligence} targets the utilization of learned world knowledge~\citep{cattell1963theory}, relying on Inductive Reasoning and Prototype Theory to abstract semantic concepts from visual variance~\citep{rosch1973natural}.

To assess 3D spatial understanding, \textbf{Visuospatial Cognition} tests the reconstruction of structural relationships, requiring the ability to group discrete visual elements into coherent, holistic structures based on Gestalt laws~\citep{wertheimer1923untersuchungen} and the decomposition of shapes into geometric primitives (geons) consistent with Recognition-by-Components Theory~\citep{Biederman1987Recognition}.

Distinct from static recognition, \textbf{Mental Simulation} demands that models act as ``simulation engines'', employing Hypothetico-Deductive Reasoning to synthesize programs and infer hidden dynamics~\citep{Battaglia2013simulation}.

Finally, \textbf{Visual Routines} evaluates efficiency in visual search, requiring the composition of elementary operations and Focused Attention to bind features and exercise inhibitory control~\citep{ullman1984visual}.\looseness-1

\cref{fig:dataset_example} depicts examples of each category, and~\cref{fig:dataset_satas} shows the data distribution of the train split of CogSense-Bench. The theoretical grounding of the five cognitive dimensions is detailed in~\Cref{appx_sec:grounding}. Detailed statistics are provided in~\cref{appx_sec:data_stats}, while~\Cref{appx_sec:data_pipeline} provides the implementation details for the data pipeline.

CogSense-Bench is designed to probe high-level visual cognition beyond recognition, including abstract rule induction, spatial structuring, mental simulation, and attention control.
These categories often require (i) \emph{explicit multi-step reasoning} that composes elementary operations, and (ii) maintaining and manipulating \emph{answer-oriented internal visual states} during inference. Motivated by these requirements, we propose Cognitive Supersensing, which couples chain-of-thought-style rationale generation with latent visual imagery prediction, enabling MLLMs to better support the simulation-like and stateful reasoning demanded by CogSense-Bench. \looseness=-1

\section{Method}
\label{sec:method}
\subsection{Preliminaries}
\label{method_subsec:preliminaries}

\begin{figure*}[t]
    \centering
    \includegraphics[width=0.99\linewidth]{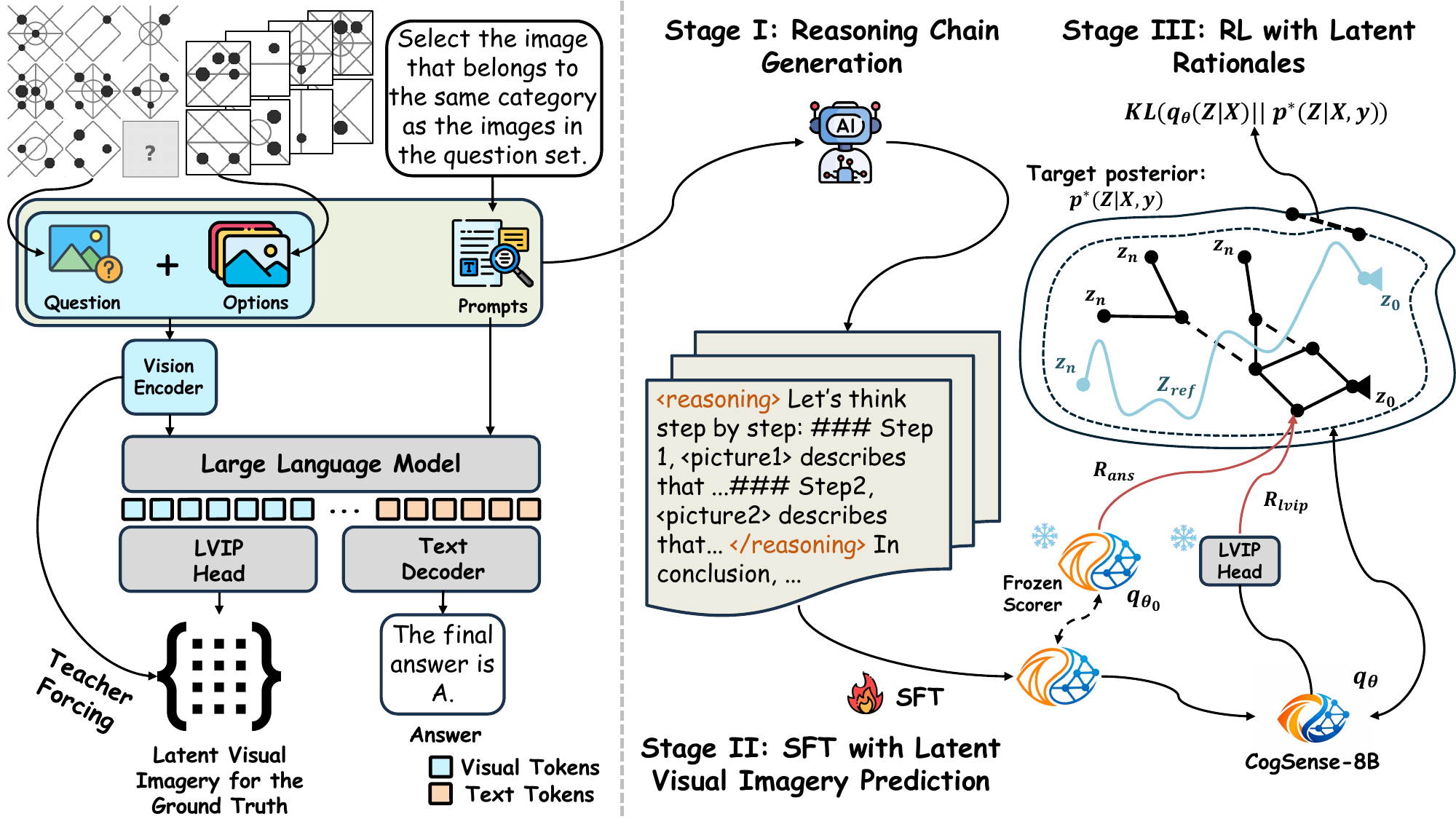}
     
    \caption{\textbf{Cognitive Supersensing framework.} \textbf{Architecture Overview (Left):} \ourmodel is a VLM that takes images and prompts as input with a text decoder to generate the answer and a Latent Visual Imagery Prediction (LVIP) head to generate a latent visual imagery of the option-image in parallel. \textbf{Method Overview (Right):} To train \ourmodel, we first synthesize teacher-guided reasoning chains (Stage I), then jointly fine-tune the model with text generation and latent visual imagery prediction objectives (Stage II), and finally optimize rationale trajectories with RL using both answer evidence and LVIP-based visual grounding (Stage III).}
    \label{fig:framework}
     
\end{figure*}

Let $\mathcal{V}$ and $\mathcal{Q}$ denote the visual input and the textual prompt, respectively, and denote the multimodal input as $X=(\mathcal{V},\mathcal{Q})$.
Our goal is to learn a multimodal model parameterized by $\theta$, which induces a conditional generation distribution $q_\theta(\cdot\mid X)$ over an output consisting of a reasoning rationale $Z$ (a token sequence) and a final answer $y$.

Specifically, $\mathcal{V}$ is processed by a pre-trained visual encoder $\Enc_{\vis}(\cdot)$ to extract visual features, denoted as
$\mathbf{V}_{\mathcal{V}}=\{\mathbf{v}_i\}_{i=1}^{T}$ with $\mathbf{v}_i\in \mathbb{R}^{P \times d_{\mathrm{vis}}}$, where $T$ is the number of images, and $P$ is the number of patches per image.
These features are mapped into the language embedding space via a projection layer $\mathcal{P}(\cdot)$, yielding projected visual tokens
$\mathbf{H}_{\mathcal{V}}=\mathcal{P}(\mathbf{V}_{\mathcal{V}})$.
Simultaneously, the prompt $\mathcal{Q}$ is tokenized and embedded using the LLM embedding layer to obtain textual tokens $\mathbf{H}_{\mathcal{Q}}$.
The LLM backbone $\Enc_{\txt}(\cdot)$ takes the concatenated token sequence $[\mathbf{H}_{\mathcal{V}},\mathbf{H}_{\mathcal{Q}}]$ as input
and produces hidden states used by (i) a text decoder to generate $(Z,y)$ autoregressively and (ii) a Latent Visual Imagery Prediction (LVIP) head to predict answer-oriented latent imagery.

Training follows a three-stage pipeline, shown in \cref{fig:framework}:
\textbf{(1)} Reasoning Chain Generation to synthesize high-quality rationales (\Cref{method_subsec:reasoning}),
\textbf{(2)} SFT with Latent Visual Imagery Prediction to jointly learn text generation and latent imagery prediction (\Cref{method_subsec:LVIP}),
and \textbf{(3)} RL with Latent Rationales to further optimize rationale sampling (\Cref{sec:stage_3}).\looseness=-1

\subsection{Stage I: Reasoning Chain Generation}
\label{method_subsec:reasoning}

To address the scarcity of high-quality cognitive training data for visual CoT reasoning, we synthesize rationales using a powerful MLLM as a teacher model, denoted as $\mathcal{M}_T$.
Given a multimodal input pair $(\mathcal{V},\mathcal{Q})$, $\mathcal{M}_T$ generates a reasoning rationale $Z$ that demonstrates the logical deduction steps, along with a predicted answer $\hat{y}$, which we then filter to ensure alignment with the ground-truth answer $y$.
Specifically, for each example, we construct a task-specific generation prompt $\mathcal{Q}_{\mathrm{gen}}$.
The teacher model is instructed to analyze the visual input and produce a step-by-step reasoning chain based on $(\mathcal{V},\mathcal{Q},\mathcal{Q}_{\mathrm{gen}})$:
\begin{equation}
    (Z,\hat{y}) \sim \mathcal{M}_T(\cdot\mid \mathcal{V},\mathcal{Q},\mathcal{Q}_{\mathrm{gen}}).
\end{equation}
We filter out generated reasoning chains that fail to reach the correct conclusion ($\hat{y} \neq y$) or contain hallucinated content.
This yields an augmented dataset $\mathcal{D}_{\mathrm{chain}}\!=\!\{(\mathcal{V}_i,\mathcal{Q}_i,Z_i,y_i)\}_{i=1}^{D}$ for SFT, where $D$ is the total number of examples. Additional details regarding $\mathcal{Q}_{\mathrm{gen}}$ are available in Appendix~\ref{appx_data_pipeline_subsec:reasoning}.

\subsection{Stage II: SFT with Latent Visual Imagery Prediction}
\label{method_subsec:LVIP}

We propose \textbf{Cognitive Supersensing} as an approach for acquiring visual cognitive capabilities. Motivated by the human \emph{constructive matching} process, we introduce an auxiliary module, Latent Visual Imagery Prediction (LVIP), to predict $\mathbf{h}_y$, the latent representation of the ground-truth answer option image. The latent visual imagery generation is synchronized with the autoregressive next-token prediction. At each decoding step of the textual CoT, the LVIP head generates a corresponding visual latent embedding based on the current context.

\noindent \textbf{LVIP Head.}
The LVIP head $g_\psi(\cdot)$ is a two-layer MLP attached to the shared LLM backbone, in addition to the text decoder. LVIP predicts the latent representation of the answer option using the backbone hidden states corresponding to visual tokens. Let $\mathbf{H}^{\mathrm{out}}_{\mathcal{V}}\in\mathbb{R}^{N\times d}$ be the sequence of hidden states for visual tokens output by $\Enc_{\txt}(\cdot)$, where $N$ is the number of visual tokens and $d$ is the hidden size of the LLM.
We extract the subset corresponding to the option images, denoted as $\mathbf{H}_{\opt}\in\mathbb{R}^{M\times d}$, where $M$ is the number of tokens corresponding to option images.
We apply average pooling over $\mathbf{H}_{\opt}$ to obtain an aggregated representation $\bar{\mathbf{H}}_{\opt}\in \mathbb{R}^{1\times d}$,
and compute the predicted latent imagery $\hat{\mathbf{h}}_y=g_\psi(\bar{\mathbf{H}}_{\opt}) \in \mathbb{R}^{1 \times d_\mathrm{vis}}$.

\noindent \textbf{Learning Objectives.}
Let $v_y$ denote the candidate option image indexed by the ground-truth answer $y$.
The supervision target for LVIP is the embedding of $v_y$ extracted by the (frozen) visual encoder and then projected into the same space as $\hat{\mathbf{h}}_y$ via an MLP, where $\mathbf{h}_y=\mathrm{MLP}(\mathbf{v}_y) \in \mathbb{R}^{1 \times d_\mathrm{vis}}.$
We optimize LVIP with an MSE loss between $\hat{\mathbf{h}}_y$ and $\mathbf{h}_y$, jointly with the standard autoregressive cross-entropy loss over the target text sequence $x=(Z,y)$ conditioned on $X=(\mathcal{V},\mathcal{Q})$:
\begin{equation}
\setlength{\abovedisplayskip}{4pt}
\setlength{\belowdisplayskip}{4pt}
\setlength{\abovedisplayshortskip}{2pt}
\setlength{\belowdisplayshortskip}{2pt}
\begin{aligned}
\mathcal{L}_{\SFT}
&=
-\sum_{t=1}^{|x|}
\log q_\theta(x_t \mid X, x_{<t}) + \beta \cdot \MSE(\hat{\mathbf{h}}_y, \mathbf{h}_y),
\end{aligned}
\end{equation}
where $\beta$ balances the two objectives. During inference, the final answer is generated by the text decoder, while the LVIP head is kept as a frozen auxiliary module to produce $\hat{\mathbf{h}}_y$ as a latent visual grounding signal that supports answer-oriented reasoning.

\subsection{Stage III: RL with Latent Rationales}
\label{sec:stage_3}
In this stage, we leverage diversity-seeking RL by performing amortized variational inference over latent rationales using a Generative Flow Network (GFlowNet)~\citep{bengio2023gflownetfoundations,lahlou2023theorycontinuousgenerativeflow}. Concretely, we refine the rationale policy $q_\theta(Z \mid X)$ to sample rationale trajectories $Z$ in proportion to an unnormalized trajectory score, rather than committing to a single deterministic chain. We define the corresponding reward-induced target posterior over rationales as
\begin{equation}
p^{\ast}(Z \mid X,y)\ \propto\ \exp\!\left(R(Z;X,y)\right),
\end{equation}
and train $q_\theta(Z\mid X)$ to approximate this posterior with a GFlowNet objective, detailed below. We define the trajectory score as a weighted combination of answer evidence and LVIP-based representation grounding:
\begin{equation}
R(Z;X,y)
=
\alpha\,R_{\mathrm{ans}}(Z;X,y)
+
\gamma\,R_{\mathrm{lvip}}(Z;X,y).
\end{equation}
Importantly, the LVIP term provides answer-oriented grounding in representation space, which complements discrete answer supervision and facilitates exploration over long rationale trajectories.

\definecolor{humangray}{RGB}{30, 90, 140}
\begin{table*}[!t]
  \centering
    \caption{\textbf{CogSense-Bench Cognitive Ability Results.} \ourmodel achieves the best results in all five dimensions and the highest Avg. (accuracy over all 1,000 test items), outperforming both closed-source MLLMs and fine-tuned (FT) open-source baselines. Best and second-best results in \textbf{bold} and \underline{underline}.}

  \label{tab:exp_cognitive}

  \setlength{\tabcolsep}{5.3pt}
  \renewcommand{\arraystretch}{1.12}
  \footnotesize

  \resizebox{0.98\linewidth}{!}{%
  \begin{tabular}{lcccccc}
    \toprule
    \rowcolor{gray!25}
    \textbf{Model} & \textbf{Fluid Intel.} & \textbf{Crystallized Intel.} & \textbf{Visuospatial Cog.} & \textbf{Mental Simu.} & \textbf{Visual Rout.} & \textbf{Avg.} \\
    \midrule

    \rowcolor{gray!15}
    \textcolor{humangray}{\textbf{Human}} & 82.7 & 91.3 & 88.5 & 97.9 & 78.7 & 88.4 \\

    \midrule
    Gemini 2.5 Flash & 23.2 & 40.2 & 31.0 & 40.0 & {45.2} & 34.9 \\
    \rowcolor{gray!8}
    GPT-o3 & 4.7 & {51.4} & 20.4 & 38.7 & 43.0 & 32.3 \\
    GPT-5.2 & 29.3 & 35.9 & {57.5} & {60.0} & 37.6 & {40.3} \\
    \rowcolor{gray!8}
    Claude Sonnet 4 & 22.5 & 31.3 & 26.5 & 58.0 & 34.4 & 32.6 \\
    Grok 4 Fast & 13.0 & 45.4 & 41.6 & 21.3 & 37.6 & 31.7 \\

    \midrule
    \rowcolor{gray!8}
    Llama-4-Scout-17B & 20.3 & 29.9 & 35.4 & 48.7 & 41.9 & 31.8 \\
    Gemma-3-27B & 18.5 & 29.3 & 39.8 & 55.3 & 43.0 & 32.7 \\
    \rowcolor{gray!8}
    Ministral-3-14B & 9.4 & 10.6 & 10.6 & 39.3 & 31.2 & 16.5 \\
    Intern-S1 & 6.5 & 20.9 & 12.4 & 17.3 & 43.0 & 17.5 \\
    \rowcolor{gray!8}
    Qwen3-VL-30B & 30.8 & 34.0 & 37.2 & 56.0 & 40.9 & 37.4 \\

    \midrule
    Qwen3-VL-8B (FT) & 51.1 & 76.6 & \underline{63.7} & 59.3 & 41.9 & 62.3 \\
    \rowcolor{gray!8}
    InternVL3.5-8B (FT) & 52.2 & 80.2 & \underline{63.7} & 59.3 & 41.9 & 63.9 \\
    Qwen3.5-9B (FT) & \underline{56.5} & \underline{89.1} & 61.9 & \underline{62.7} & \underline{46.2} & \underline{69.1} \\

    \midrule
    \ourmodel (Ours) &~~~~~~\textbf{63.8}\red{7.3} &~~~~~~\textbf{91.0}\red{1.9} &~~~~~~\textbf{69.0}\red{7.1} &~~~~~~\textbf{68.0}\red{5.3} &~~~~~~\textbf{50.5}\red{4.3} &~~~~~~\textbf{73.8}\red{4.7} \\
    \bottomrule
  \end{tabular}%
  }
\end{table*}

\noindent \textbf{Token-Wise Marginal Reward Estimation.}
Let $Z=\left(z_1, \ldots, z_L, \top\right)$ be an autoregressive rationale, with $\top$ denoting end of sequence. We define the answer-evidence component using a frozen scorer to avoid a moving-target reward:
\begin{equation}
R_{\mathrm{ans}}(Z;X,y)
=
\log q_{\theta_0}\!\left(y \mid X,Z\right),
\end{equation}
where $q_{\theta_0}$ is a frozen scorer (a fixed copy of the supervised model). To incorporate LVIP grounding during RL while keeping it stationary, we keep the LVIP head $g_\psi$ frozen and compute a representation-level reward that depends on $Z$ through the backbone conditioning on $[X;Z]$. Let $v_y$ be the option image indexed by the ground-truth answer $y$, and let $\mathbf{h}_y=\mathrm{MLP}(\mathbf{v}_y)$ be its embedding from a frozen visual encoder followed by an MLP. Let $\bar{\mathbf{H}}_{\opt}(X,Z;\theta)$ denote the average-pooled final-layer hidden state of the option-image visual tokens produced by the backbone when conditioned on $[X;Z]$. We define: \looseness=-1
\begin{equation}
R_{\mathrm{lvip}}(Z;X,y)
=
-\left\|g_\psi\!\big(\bar{\mathbf{H}}_{\opt}(X,Z;\theta)\big)-\mathbf{h}_y\right\|_2^2.
\end{equation}
To provide prefix-level training signals required by flow-based objectives, we define the prefix state at step $t$ as $\tau_t=(z_{1:t},\top)$, where $\top$ indicates the availability of a termination action rather than an emitted token. Instead of evaluating the scorer at every token, we compute $\widetilde R(\tau_t)$ only at sparse \emph{anchor} indices with stride $\lambda$ and linearly interpolate within each segment. Specifically, let anchors be $t\in\{0,\lambda,2\lambda,\ldots\}$ with $t<L$, and define $t^{+}=\min(t+\lambda,L)$. For any integer $i\in\{0,1,\ldots,t^{+}-t\}$,
\begin{equation}
\widetilde{R}(\tau_{t+i})
=
\widetilde{R}(\tau_t)
+\frac{i}{t^{+}-t}\Big(\widetilde{R}(\tau_{t^{+}})-\widetilde{R}(\tau_t)\Big).
\end{equation}
At anchor positions, we set $\widetilde R(\tau_t)=R(\tau_t;X,y)$ with
\begin{equation}
\begin{aligned}
R(\tau_t;X,y)
&=
\alpha\,\log q_{\theta_0}\!\left(y \mid X, z_{1:t}\right) +\gamma\,R_{\mathrm{lvip}}(\tau_t;X,y),
\end{aligned}
\end{equation}
where $R_{\mathrm{lvip}}(\tau_t;X,y)$ is computed analogously by conditioning the backbone on $[X;z_{1:t}]$. We employ a standard SubTB loss~\citep{madan2023learninggflownetspartialepisodes} to train $q_\theta(Z \mid X)$ toward the reward-induced target distribution. \looseness=-1

\noindent \textbf{Reference-Guided GFlowNet Fine-tuning.}
To reduce variance from low-quality samples, we anchor exploration with a reference rationale $Z_{\mathrm{ref}}$. For each $X$, we sample $m$ candidates $\left\{Z_i\right\}_{i=1}^m \sim q_\theta(\cdot \mid X)$ and keep only those that meet a relative evidence threshold:
\begin{equation}
\setlength{\abovedisplayskip}{4pt}
\setlength{\belowdisplayskip}{4pt}
\setlength{\abovedisplayshortskip}{2pt}
\setlength{\belowdisplayshortskip}{2pt}
\mathbb I(Z_i)=\mathbf 1\!\left[\,R(Z_i;X,y)\;\ge\;R(Z_{\mathrm{ref}};X,y)+\log\delta_s\,\right],
\end{equation}
where $\delta_s\in(0,1]$ and $s$ is the index of the current training step (so $\log\delta_s\le 0$ controls the allowed slack in log-space relative to the reference). We optimize SubTB only on accepted trajectories:
\begin{equation}
\setlength{\abovedisplayskip}{4pt}
\setlength{\belowdisplayskip}{4pt}
\setlength{\abovedisplayshortskip}{2pt}
\setlength{\belowdisplayshortskip}{2pt}
\mathcal L(\theta)=\sum_{i=1}^{m}\mathbb I(Z_i)\cdot \mathcal L_{\mathrm{SubTB}}(Z_i;\theta),
\end{equation}
where $\mathcal L_{\mathrm{SubTB}}$ is instantiated with the densified prefix scores $\widetilde R$.

\noindent \textbf{Bayesian Posterior over Latent Rationales.}
At inference time, we treat $Z$ as latent and aggregate evidence across multiple sampled rationales. We sample $n$ rationales $\{Z_i\}_{i=1}^{n}\sim q_\theta(Z\mid X)$, decode an answer $y_i$ conditioned on $(X,Z_i)$, and compute a length-normalized evidence score with the frozen scorer: \looseness=-1
\begin{equation}
S_i=\frac{1}{|Z_i|+|y_i|}\log q_{\theta_0}(y_i\mid X,Z_i).
\end{equation}
We then output $\hat y = y_{i^{\ast}}$, where $i^{\ast}=\operatorname*{arg\,max}_i S_i$, which serves as a simple MAP-style selection over sampled latent rationales and reduces sensitivity to any single brittle chain under reasoning ambiguity.

\begin{figure*}[!t]
    \centering
    \includegraphics[width=0.99\linewidth]{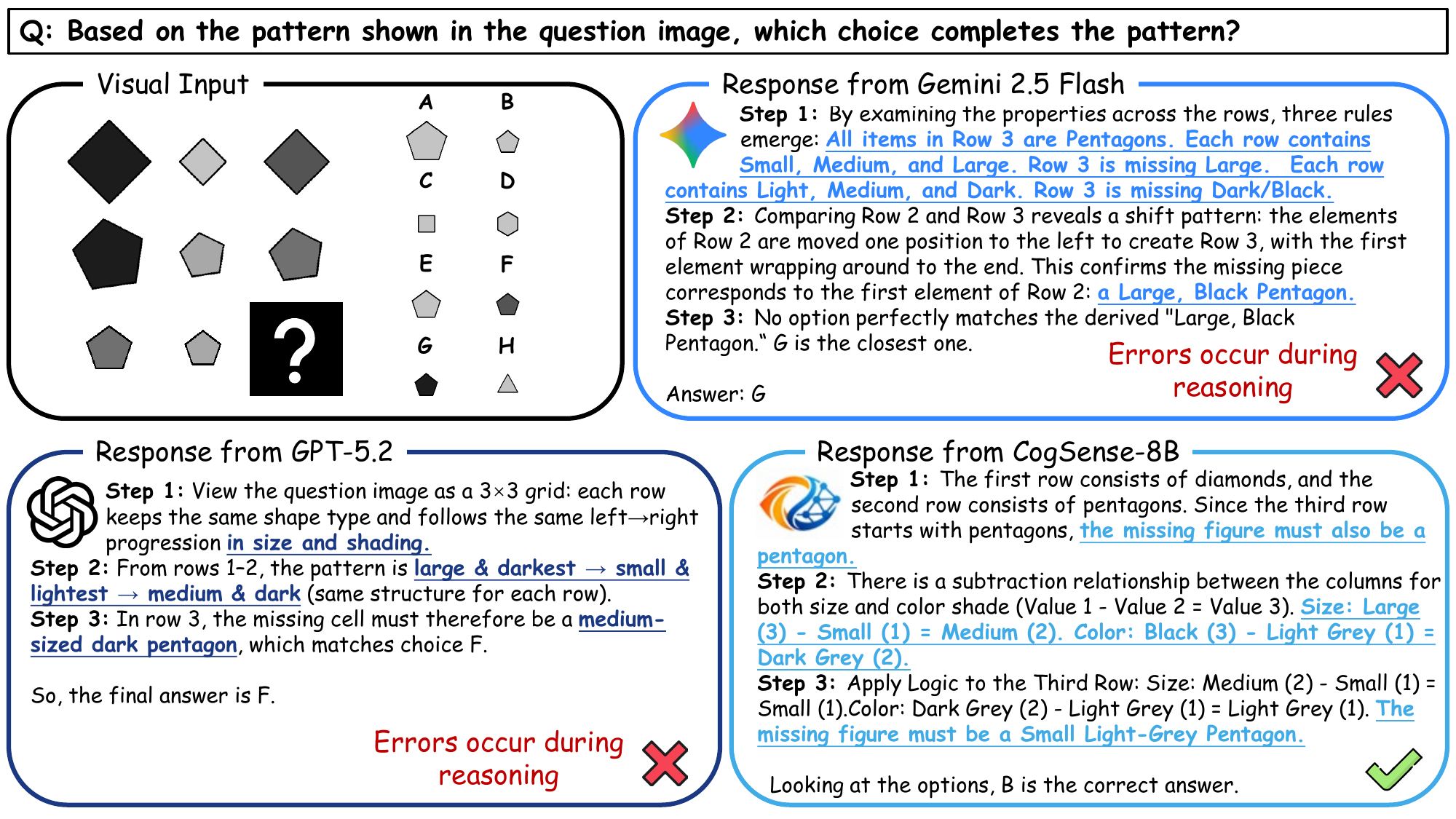}
     
    \caption{\textbf{Qualitative Example of Visual Cognition Reasoning Across Models.} We \underline{underline} decisive sentences in the reasoning chain. \ourmodel demonstrates a coherent, multi-step logical chain that closely matches the ground truth, while other models exhibit less precise or less interpretable reasoning paths.}
    \label{fig:qualitative_example}
\end{figure*}

\section{Experiments}
\label{sec:exp}

\noindent \textbf{Baselines and Metrics.} We compare \ourmodel with several mainstream open-source and closed-source MLLMs and VLMs with strong multimodal performance: Gemini 2.5 Flash~\citep{comanici2025gemini25pushingfrontier}, GPT-o3~\citep{openai2025o3}, GPT-5.2~\citep{openai2025gpt5.2}, Claude Sonnet 4~\citep{anthropic_claude4_system_card_2025}, Grok 4 Fast~\citep{xai2025grok4fast_modelcard}, Llama-4-Scout-17B~\citep{meta2025llama4}, Gemma-3-27B~\citep{gemmateam2025gemma3technicalreport}, Ministral-3-14B~\citep{ministral}, Intern-S1~\citep{bai2025interns1scientificmultimodalfoundation}, and Qwen3-VL-30B~\citep{bai2025qwen3vltechnicalreport}. To provide a fair comparison, we also fine-tune Qwen3-VL-8B~\citep{bai2025qwen3vltechnicalreport}, InternVL3.5-8B~\citep{wang2025internvl3}, and Qwen3.5-9B~\citep{qwen35blog} on our train split with the same settings and then evaluate them. We use accuracy as the evaluation metric. We further recruited 20 participants to answer 100 questions selected from CogSense-Bench via stratified sampling, and report their accuracy in Table~\ref{tab:exp_cognitive} as a human baseline. Detailed information about the human study can be found in~\cref{appx_sec:human_study}. We also compare \ourmodel with ROSS~\citep{wang2024reconstructive}, which reconstructs question images for self-supervised learning. A detailed comparison can be found in~\Cref{appx_sec:comparision}. \looseness=-1

\noindent \textbf{Implementation Details.} All experiments are conducted with 8 NVIDIA H200 GPUs. We use Qwen3-VL-8B as the backbone to train \ourmodel. We use AdamW as the optimizer with a learning rate of $\eta=10^{-5}$ and a weight decay of $10^{-5}$. We use GPT-5~\citep{singh2025openai} as the teacher model in Stage I. \looseness=-1

\subsection{Cognitive Ability Results}

\noindent \textbf{Quantitative Results.}  As~\Cref{tab:exp_cognitive} illustrates, \ourmodel narrows the performance gap relative to existing MLLMs, suggesting that it effectively uses visual imagery to support cognitive reasoning. Despite being an 8B model, it achieves SoTA performance on all tasks and delivers the strongest overall accuracy (\textbf{73.8}), surpassing GPT-5.2 by \textbf{+33.5}. It also outperforms all fine-tuned baselines, confirming that our strong performance stems from our methodological contributions rather than mere dataset exposure. Notably, for highly symbolic tasks like Visual Routines, which are image-only and are difficult even for humans to solve, the model still outperforms the baselines, indicating that the average pooling operation in LVIP does not discard fine-grained visual information; rather, LVIP acts as a guiding signal. A statistical analysis of the experimental results is provided in~\Cref{appx_sec:analysis}.\looseness=-1

\noindent \textbf{Qualitative Comparison.} \Cref{fig:qualitative_example} shows a qualitative comparison between \ourmodel and mainstream models, demonstrating that Gemini 2.5 Flash and GPT-5.2 both fail to extract the correct underlying rule of the pattern, leading to incorrect answers. In contrast, \ourmodel demonstrates high-quality reasoning with a concise and clear expression of the underlying pattern. This comparison suggests that \ourmodel more effectively captures the underlying visual regularities required for abstract pattern reasoning. Additional qualitative examples can be found in~\cref{appx_sec:qualitative_examples}.

\subsection{General Ability Results}
We also evaluate the general vision-language understanding and spatial reasoning abilities of \ourmodel. As shown in Table~\ref{tab:exp_general}, we evaluate on vision-language datasets such as GQA~\citep{hudson2019gqanewdatasetrealworld}, ScienceQA~\citep{lu2022learnexplainmultimodalreasoning}, RealWorldQA~\citep{grokv2024}, BLINK~\citep{fu2024blinkmultimodallargelanguage}, and HallusionBench~\citep{guan2024hallusionbenchadvancediagnosticsuite}, and spatial-related benchmarks such as MMSI-Bench~\citep{yang2025mmsi}, ViewSpatial-Bench~\citep{li2025viewspatial}, and ERQA~\citep{team2025gemini}. \ourmodel maintains robust performance on these general benchmarks, comparable to the base model. This balance suggests that \ourmethod's visual cognitive gains do not stem from overfitting on the training data; instead, our method injects high-level visual cognitive knowledge while preserving the foundational knowledge and instruction-following capabilities inherent in the pre-trained backbone. We attribute the performance decline on the \textbf{OCR-related} benchmark, HallusionBench, to a natural post-training shift in which the model trades some specialized text-extraction proficiency for stronger pure visual-spatial logic. Additional general-ability evaluations are in~\Cref{appx_sec:general}.

\begin{table*}[!t]
  \centering
  \caption{\textbf{General Ability Results.} Performance comparison in general tasks with \ourmodel and the base model. \ourmodel demonstrates similar general ability performance to the base model.}

  \label{tab:exp_general}

  \setlength{\tabcolsep}{5.3pt}
  \renewcommand\arraystretch{1.15}

  \rowcolors{2}{gray!10}{white}

  \footnotesize
  \resizebox{\linewidth}{!}{%
  \begin{tabular}{lcccccccc}
    \toprule
    \rowcolor{gray!25}
    \textbf{Model} & \textbf{GQA} & \textbf{ScienceQA} & \textbf{RealWorldQA} & \textbf{BLINK} & \textbf{HallusionBench} & \textbf{MMSI-Bench} & \textbf{ViewSpatial-Bench} & \textbf{ERQA} \\
    \midrule
    Qwen3-VL-8B (Base) & 71.4 & \textbf{92.6} & 71.5 & 64.7 & \textbf{61.1} & 28.0 & 38.6 & 46.8 \\
    \ourmodel (Ours) & \textbf{71.8} & \textbf{92.6} & \textbf{71.9} & \textbf{65.3} & 60.5 & \textbf{32.4} & \textbf{40.8} & \textbf{48.8} \\
    \bottomrule
  \end{tabular}}
\end{table*}

\begin{table*}[!t]
    \centering
    \caption{\textbf{Module Ablation Study Results.} We compare the base model and the following variants: SFT \textit{w/o} LVIP, SFT \textit{w/} LVIP, SFT \textit{w/o} LVIP + GRPO, SFT \textit{w/} LVIP + GRPO, SFT \textit{w/o} LVIP + GRPO + Latent Rationales (LR), SFT \textit{w/} LVIP + GRPO + LR, SFT \textit{w/o} LVIP + LR, and \ourmodel. We \textbf{bold} the best results and \underline{underline} the runner-ups. \looseness=-1}

    \label{tab:exp_ablation}
    \setlength{\tabcolsep}{6pt}
    \renewcommand\arraystretch{1.15}
    \rowcolors{2}{gray!10}{white}
    \resizebox{\linewidth}{!}{%
      \begin{tabular}{lcccccc}
        \toprule
        \rowcolor{gray!25}
        \textbf{Variant} & \textbf{Fluid Intel.} & \textbf{Crystallized Intel.} & \textbf{Visuospatial Cog.} & \textbf{Mental Simu.} & \textbf{Visual Rout.} & \textbf{Avg.} \\
        \midrule
        Qwen3-VL-8B (Base) & 31.2 & 34.8 & 31.0 & 45.3 & 40.9 & 35.5 \\
        Qwen3-VL-8B SFT \textit{w/o} LVIP  & 51.1 & 76.6 & {63.7} & 59.3 & 41.9 & 62.3 \\
        Qwen3-VL-8B SFT \textit{w/} LVIP  & {55.4} & {88.6} & 61.1 & {61.3} & {44.1} & {68.1} \\
        Qwen3-VL-8B SFT \textit{w/o} LVIP + GRPO  & 55.8 & 79.9 & 63.7 & 63.3 & 43.0 & 65.5 \\
        Qwen3-VL-8B SFT \textit{w/} LVIP + GRPO &
        59.1 & 89.9 & 64.6 & 65.3 & 46.2 & 70.8 \\
        Qwen3-VL-8B SFT \textit{w/o} LVIP + GRPO + LR & 56.5 & 80.4 & 64.6 & 64.0 & 44.1 & 66.2 \\
        Qwen3-VL-8B SFT \textit{w/} LVIP + GRPO + LR & \underline{59.8} & \underline{90.2} & \underline{65.5} & \underline{66.7} & \underline{47.3} & \underline{71.5} \\
        Qwen3-VL-8B SFT \textit{w/o} LVIP + LR & 57.2 & 80.7 & 65.5 & 65.3 & 45.2 & 66.9 \\
        \ourmodel (Ours) & \textbf{63.8} & \textbf{91.0} & \textbf{69.0} & \textbf{68.0} & \textbf{50.5} & \textbf{73.8} \\

        \bottomrule
      \end{tabular}
    }
\end{table*}

\subsection{Ablation Studies}
\noindent \textbf{Module Ablation.} As summarized in Table~\ref{tab:exp_ablation}, to isolate the contributions of LVIP and our proposed reinforcement learning mechanism, we perform a hierarchical evaluation across various variants. Compared to the vanilla Qwen3-VL-8B, applying standard SFT substantially improves the average accuracy from 35.5 to 62.3. The integration of LVIP introduces a further improvement, raising the average accuracy to \textbf{68.0}. This suggests that LVIP helps the model align semantic reasoning and abstract world modeling, enabling the model to gain initial ``supersensing'' capabilities. To validate the necessity of our specific RL design, we introduce GRPO as a reinforcement learning baseline. Although applying GRPO to the SFT models (\textit{w/o} or \textit{w/} LVIP) yields observable gains, achieving improvements of \textbf{+3.2} and \textbf{+2.8} compared to models without GRPO, it serves primarily to refine the policy within a standard optimization scope. GRPO plus Latent Rationales further improves performance to \textbf{66.2} and \textbf{71.5} compared to GRPO only, suggesting that the two RL objectives are complementary. Finally, \ourmodel achieves the best average performance of $\textbf{73.8}$, outperforming the strongest baseline (SFT \textit{w/} LVIP + GRPO + Latent Rationales) by a clear margin. This result highlights that our elaborately designed RL method is not merely an optimization trick but a specialized reasoning refiner that is more effective than general GRPO in leveraging the visual supersensing established by LVIP. \looseness=-1

\noindent \textbf{Latent Visual Imagery Learning.} To verify that LVIP learns meaningful intermediate visual states rather than answer-conditioned shortcuts, we perform three experiments. \textbf{(i) Noise Injection~\citep{li2026dynamics}:} Perturbing the latent state at Step 30 increases the MSE loss between intermediate latent visual imagery at multiple steps of the CoT and the final ground-truth latent visual imagery (\Cref{tab:exp_noise}), demonstrating that the uncorrupted intermediate state is vital for reasoning. \textbf{(ii) Latent Patching~\citep{liang2026latent}:} Swapping latent states between correct and incorrect runs shifts the rationale and answer probability toward the patched state (\Cref{tab:exp_patching}), showing that these states carry causal information that steers inference. \textbf{(iii) Linear Probing~\citep{alain2016understanding}:} A linear classifier trained on intermediate latent visual imagery at multiple steps of the CoT shows that accuracy grows progressively (\Cref{tab:exp_probing}). This confirms that visual logic is constructed step-by-step, ruling out trivial answer-conditioned shortcuts. Detailed settings, more experiments, and discussions of latent visual imagery learning can be found in~\Cref{appx_sec:lvip}. \looseness=-1
\begin{table*}[!t]
    \centering
    \caption{\textbf{Quantitative Analysis of Latent Visual Imagery Learning.} The three sub-tables report the MSE loss after noise injection, probability changes from latent patching, and linear probing accuracy, respectively.}
    \label{tab:exp_shortcut}

    \begin{subtable}{0.44\linewidth}
        \centering
        \setlength{\tabcolsep}{3pt}
        \renewcommand\arraystretch{1.15}
        \rowcolors{2}{gray!10}{white}
        \caption{\textbf{Noise Injection.}}

        \label{tab:exp_noise}
        \resizebox{\linewidth}{!}{%
            \begin{tabular}{lccccc}
              \toprule
              \rowcolor{gray!25}
              \textbf{Intervention} & \textbf{Step 0} & \textbf{Step 10} & \textbf{Step 20} & \textbf{Step 30} & \textbf{Step 40} \\
              \midrule
              Add Gaussian Noise @ Step30 & 0.492 & 0.368 & 0.215 & 0.374 & 0.418 \\
              \ourmodel \textit{w/o} Noise & 0.492 & 0.368 & 0.215 & 0.184 & 0.158 \\
              \bottomrule
            \end{tabular}
        }
    \end{subtable}
    \hfill
    \begin{subtable}{0.19\linewidth}
        \centering
        \setlength{\tabcolsep}{3pt}
        \renewcommand\arraystretch{1.15}
        \rowcolors{2}{gray!10}{white}
        \subcaption{\textbf{Latent Patching.}}

        \label{tab:exp_patching}
        \resizebox{\linewidth}{!}{%
            \begin{tabular}{lc}
              \toprule
              \rowcolor{gray!25}
              \textbf{Patch Type} & \textbf{$\Delta p$} \\
              \midrule
              Correct$\rightarrow$Incorrect Patch & $-0.43$ \\
              Incorrect$\rightarrow$Correct Patch & $+0.22$ \\
              \bottomrule
            \end{tabular}
        }
    \end{subtable}
    \hfill
    \begin{subtable}{0.34\linewidth}
        \centering
        \setlength{\tabcolsep}{3pt}
        \renewcommand\arraystretch{1.15}
        \rowcolors{2}{gray!10}{white}
        \caption{\textbf{Linear Probing.}}

        \label{tab:exp_probing}
        \resizebox{\linewidth}{!}{%
            \begin{tabular}{cccccc}
              \toprule
              \rowcolor{gray!25}
              & \textbf{Step 0} & \textbf{Step 10} & \textbf{Step 20} & \textbf{Step 30} & \textbf{Step 40} \\
              \midrule
              \textbf{Acc.} & 0.173 & 0.237 & 0.394 & 0.552 & 0.621 \\
              \bottomrule
            \end{tabular}
        }
    \end{subtable}
\end{table*}

\FloatBarrier
\Needspace{0.7\textheight}
\begin{wrapfigure}[30]{r}{0.58\textwidth}

	\centering
	\includegraphics[width=0.99\linewidth]{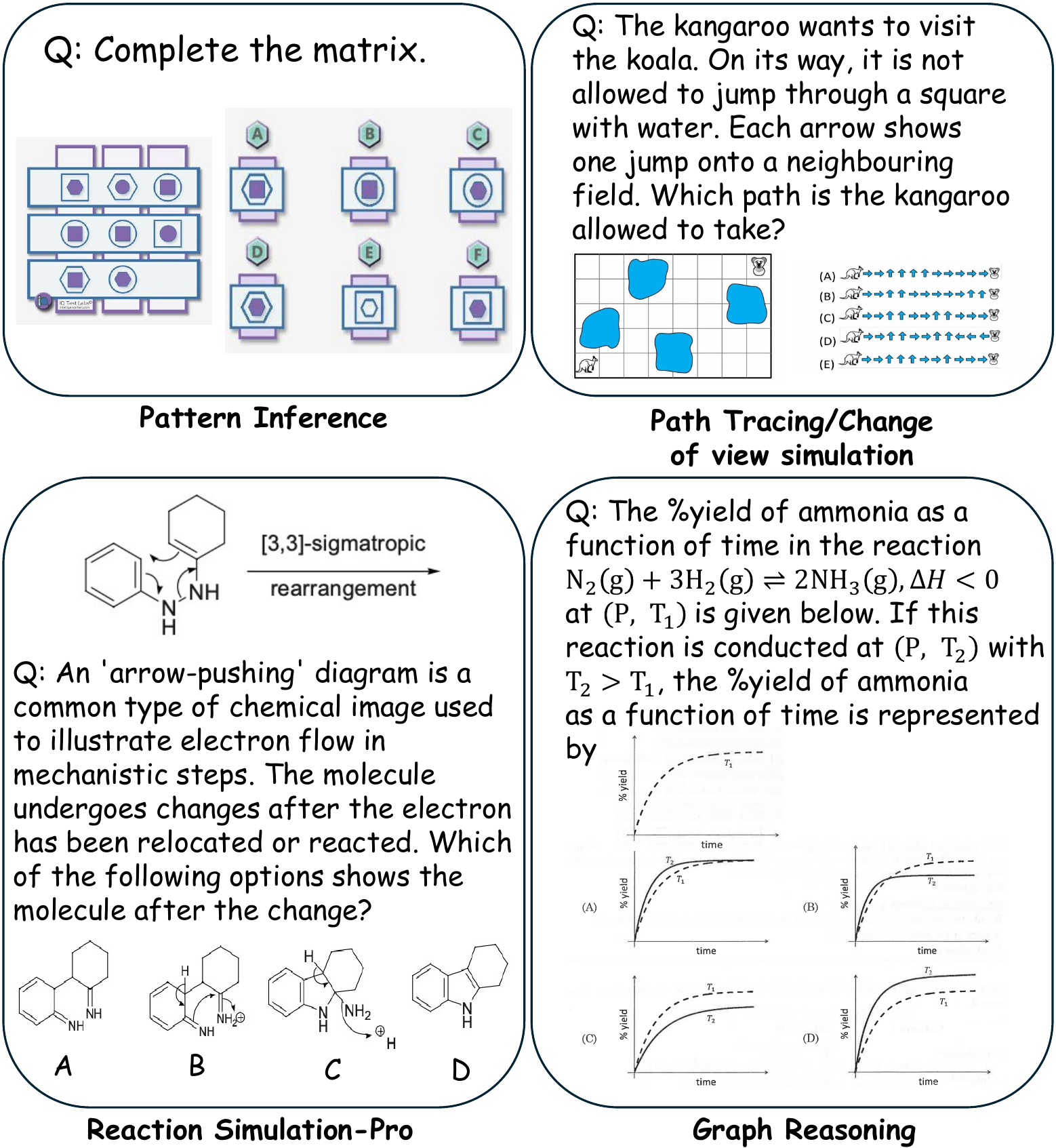}

	\caption{\textbf{EMMA Benchmark Sample Problems.}}
	\label{fig:emma_examples}

	\centering
	\captionof{table}{\textbf{Out-of-Domain Evaluation}. Generalization ability of CogSense-8B against the backbone on the EMMA benchmark.}
	\label{tab:exp_out}
	\setlength{\tabcolsep}{6pt}
	\renewcommand\arraystretch{1.15}
	\rowcolors{2}{gray!10}{white}
	\resizebox{0.95\linewidth}{!}{%
		\begin{tabular}{lcc}
			\toprule
			\rowcolor{gray!25}
			\textbf{Model} & \textbf{Chemistry} & \textbf{Mathematics} \\
			\midrule
			Qwen3-VL-8B (base) & 39.2 & 26.0 \\
			CogSense-8B (Ours) & ~~~~~~\textbf{45.4}\red{6.2} & ~~~~~~\textbf{34.8}\red{8.8} \\
			\bottomrule
		\end{tabular}
	}

\end{wrapfigure}
\noindent \textbf{Additional Ablation Studies.} We also conduct ablation studies on CoT (\Cref{appx_sec:lvip}), prefix sensitivity (\Cref{appx_sec:lvip}), the number of sampled rationales (\Cref{appx_sec:ablation}), sampling strategies (\Cref{appx_sec:strategy}), teacher models (\Cref{appx_sec:teacher}), and backbone models (\Cref{appx_sec:backbone}). \looseness=-1

\subsection{Out-of-Domain Evaluation}

To further verify the generalization capability of \ourmodel, we extend our evaluation to out-of-domain scenarios using the Chemistry and Math subsets of the EMMA benchmark~\citep{hao2025mllmsreasonmultimodalityemma}, where both the questions and the options contain images. Selected EMMA examples are shown in~\cref{fig:emma_examples}. As shown in Table~\ref{tab:exp_out}, \ourmodel achieves substantial gains of $\textbf{+6.2}$ and $\textbf{+8.8}$ on the Chemistry and Math subsets, respectively, confirming that our method learns generalized visual cognition patterns rather than overfitting to specific training data. We also evaluate \ourmodel on OOD open-ended VQA benchmarks in~\Cref{appx_sec:ood}. \looseness-1

\section{Conclusion}
\label{sec:conclusion}

We propose Cognitive Supersensing, a paradigm designed to bridge the gap between perceptual processing and complex cognitive reasoning in MLLMs. By integrating a Latent Visual Imagery Prediction (LVIP) head and employing a training strategy that combines SFT and RL with Latent Rationales, we enable \ourmodel to simulate internal visual imagery aligned with semantic reasoning chains. We further introduce CogSense-Bench, a comprehensive evaluation suite targeting five dimensions of visual cognition. Our experiments demonstrate that \ourmodel achieves superior performance compared to SoTA MLLM baselines on these cognitive tasks. These results indicate that modeling the interaction between visual simulation and logical deduction is an effective direction for advancing the reasoning abilities of multimodal systems.

\section*{Acknowledgments}
We acknowledge the support of the Health Care Engineering Systems Center within the Coordinated Science Lab at UIUC. This research was partially supported by the NSF CAREER Award \#2542328, the U.S. Defense Advanced Research Projects Agency (DARPA) under award HR001125C0303, and the U.S. Army under contract W5170125CA160. The views and conclusions contained herein are those of the authors and should not be interpreted as necessarily representing the official policies, either expressed or implied, of NSF, DARPA, the U.S. Army, or the U.S. Government. The U.S. Government is authorized to reproduce and distribute reprints for governmental purposes notwithstanding any copyright annotation therein. Portions of this work were also supported by Google PhD Fellowship, PediaMed AI and NemX Labs' innovation fund for computational resource and travel.

\bibliographystyle{bibstyle}
\bibliography{main}

\clearpage
\appendix
\section*{Appendix}\label{sec:appendix}
\crefalias{section}{appendix}
\renewcommand\thefigure{\Alph{section}\arabic{figure}}
\renewcommand\thetable{\Alph{section}\arabic{table}}

\section{Related Works}
\label{appx_sec:related}
\setcounter{figure}{0}
\setcounter{table}{0}

\noindent \textbf{Bongard Problems.} The Bongard Problems (BPs) were introduced by M. M. Bongard~\citep{bongard1968recognition, bongard1970pattern} as a novel challenge to machine vision. A typical BP includes two sets, positive and negative, with each side consisting of six images that share a common pattern. It requires a system to induce the logical rule that distinguishes a set of positive examples from negative ones~\citep{bongard1968recognition, bongard1970pattern}. This task evaluates core properties of human cognition: identifying the underlying rule that differentiates the two sides and articulating it in natural language~\citep{malk2025reasoninglimitationsmultimodallarge}. To modernize this benchmark for deep learning, the Bongard-LOGO dataset was introduced, transforming the concept learning challenge into a few-shot binary classification problem~\citep{nie2020bongard}. Recent works have expanded the field of BPs to include real-world images, such as Bongard-HOI~\citep{jiang2022bongard}, Bongard-OpenWorld~\citep{wu2024bongardopenworld}, Bongard-RWR~\citep{malk2025reasoninglimitationsmultimodallarge}, and Bongard-RWR+~\citep{pawlonka2025bongardrwrrealworldrepresentationsfinegrained}. These works expand the range of presented objects, attributes, and relationships, illustrating human-object interactions and incorporating free-form concepts in the real world, thus increasing the diversity of featured scenes.

\noindent \textbf{Matrix Reasoning.} Matrix reasoning tasks play a critical role in assessing human intelligence, particularly in relation to visual cognition and working memory~\citep{salthouse1993influence, jaeggi2010relationship, fleuret2011comparing}. These tasks are widely used in psychology through Raven's Progressive Matrices (RPMs)~\citep{raven1936mental, raven1998raven, raven2003raven}, the Wechsler Intelligence Scale for Children (WISC)~\citep{wechsler1949wechsler, kaufman2015intelligent}, and the Wechsler Adult Intelligence Scale \mbox{(WAIS-IV)}~\citep{wechsler2008wais} to evaluate human fluid intelligence and abstract visual reasoning. Early works, such as PGM~\citep{barrett2018measuring} and RAVEN~\citep{zhang2019raven}, were proposed to test whether neural networks can learn abstract reasoning and to lift machine intelligence by associating vision with structural, relational, and analogical reasoning in a hierarchical representation. Subsequent works further demonstrated that deep learning models can be trained to solve simple matrix reasoning tasks~\citep{malkinski2025deep, xu2023abstract, malkinski2024one}. Several datasets and benchmarks have also been introduced to enlarge this field, such as I-RAVEN~\citep{hu2022stratifiedruleawarenetworkabstract}, RAVEN-FAIR~\citep{benny2021scale}, and CVR~\citep{zerroug2022benchmark}. However, these studies tend to overlook the ability of humans to solve such puzzles in a zero-shot manner, without the need for explicit training on extensive datasets. Therefore, some useful zero-shot visual reasoning inference datasets such as RAVEN-IQ~\citep{huang2023languageneedaligningperception} and Visual Reasoning Benchmark~\citep{zhang2024farintelligentvisualdeductive} have been proposed to deal with the issue. The latest MaRs-VQA~\citep{cao2025visualcognitiongaphumans} addresses the limitations of previous works, namely the lack of rigorous human experiments as a reference and the reliance on relatively small datasets without psychometric validation.\looseness-1

\noindent \textbf{Abstraction and Reasoning Corpus.} Abstraction and Reasoning Corpus for Artificial General Intelligence (ARC-AGI) is a novel general AI benchmark designed to measure a human-like form of general fluid intelligence and enable fair general intelligence comparisons between AI systems and humans~\citep{chollet2019measure}. Its measurement standard emphasizes skill-acquisition efficiency and out-of-domain generalization over mere task performance. The tasks require systems to solve grid-based visual program synthesis problems by abstracting the underlying rules from only a minimal set of input-output examples (typically three to five pairs). A robust estimate of human performance~\citep{legris2024harcrobustestimatehuman} on the full ARC public evaluation set places the average human accuracy at 64.2\%, far exceeding that of current state-of-the-art AI methods. Its updated version, ARC-AGI-2~\citep{chollet2025arcagi2newchallengefrontier}, is more challenging. Even the most powerful models at the time of its release, such as OpenAI's o3~\citep{openai2025o3}, achieve only negligible performance on the newly released, adversarially selected ARC-AGI-2 challenge. This reinforces that modern frontier reasoning models fundamentally lack the structural inductive biases that are necessary for human-like skill-acquisition and generalization.

\noindent \textbf{VLMs for Vision Reasoning.} Visual recognition-related Vision Language Model (VLM) studies have made great progress since the development of CLIP~\citep{radford2021learningtransferablevisualmodels}. Prior work has shown that VLMs are capable of addressing vision reasoning tasks~\citep{zellers2019recognitioncognitionvisualcommonsense, bordes2024introductionvisionlanguagemodeling,shen2026fine,liu2026palm, li2026vis, li2025metal, li2024control, zhang2026linkedout,nguyen2025calico,wahed2024prima,li2026counterfactual,zhou2026tiny}. VLMs are the dominant architecture for bridging perception and language, typically integrating a visual encoder (\eg CLIP) with pretrained LLMs via a cross-modal connector to align visual features with text space~\citep{liu2024visual,cao2024maplm, zhang2024mmllmsrecentadvancesmultimodal, shao2024visualcotadvancingmultimodal, fu2024blinkmultimodallargelanguage, zhang2025vqtoken, zhang2026dive}. The training pipeline for VLMs often involves pre-training for modality alignment, followed by instruction tuning or Supervised Fine-Tuning (SFT) to enhance general multimodal capabilities~\citep{liu2024visual, ye2024mplugowlmodularizationempowerslarge}. Models such as Qwen-VL~\citep{bai2023qwen, bai2025qwen25vltechnicalreport, bai2025qwen3vltechnicalreport}, LLaVA~\citep{liu2024visual, liu2024improved}, MiniGPT-4~\citep{zhu2023minigpt4enhancingvisionlanguageunderstanding}, InstructBLIP~\citep{dai2023instructblipgeneralpurposevisionlanguagemodels}, and CogVLM~\citep{wang2024cogvlmvisualexpertpretrained} highlight the significance of employing high-quality visual instruction tuning data. However, current VLMs still face challenges in adapting to high-resolution~\citep{carvalho2025efficientarchitectureshighresolution, li2025inferenceoptimalvlmsneed} and visually complex images in vision reasoning. This is because most current vision reasoning approaches are primarily text-level, with the LLM exploring textual tokens while the visual input remains static~\citep{li2025latentvisualreasoning}. This absence of a dynamic visual search mechanism limits the model's ability to acquire fine-grained visual cues selectively. Other limitations stem from few-shot reasoning~\citep{guo2023how, li2026enhancing}, compositional understanding~\citep{yuksekgonul2023visionlanguagemodelsbehavelike, che2026mllms}, the constrained visual grounding capabilities inherent in CLIP~\citep{Tong_2024_CVPR, zhang2023camouflaged}, the brittleness of textual gradient-based optimization~\citep{xu2026steve}, and the challenge of dynamically integrating multimodal evidence~\citep{shen2026cogniroute, li2026long, li2026joint, li2026pearl}. Our work addresses these problems, enabling high-performing vision reasoning via step-by-step latent visual imagery prediction.\looseness-1

\noindent \textbf{Latent Visual Reasoning.} While standard VLMs rely on static visual encoding, a new paradigm of \textit{Latent Visual Reasoning} has emerged, shifting the focus from passive perception to active, internal simulation. This approach draws inspiration from World Models~\citep{zhu2025surveylatentreasoning, sun2025latentchainofthoughtvisualreasoning, zhang2026thinkjepa, zhang2026survey, li2026trimming}, enabling systems to ``think'' in a compressed latent space before generating a response. Recent advancements extend this paradigm to predict dynamic changes via temporal latent reconstruction~\citep{shen2026chronovision, zhang2024oostraj, zhang2023layout} and embed reasoning within goal-directed egocentric simulators~\citep{shen2026egoforge, li2026motormind}. Rather than mapping pixels directly to text, these models perform intermediate reasoning steps, akin to System-2 cognition, by predicting future states or manipulating visual abstractions~\citep{zhu2025surveylatentreasoning, sun2025latentchainofthoughtvisualreasoning, jia2026learning}. For instance, recent frameworks introduce ``visual scratchpads'' or latent tokens that allow the model to sketch out reasoning traces implicitly~\citep{zhang2025latentsketchpadsketchingvisual, li2025imaginereasoningspacemultimodal}. Methods such as Mirage~\citep{yang2025machinementalimageryempower} and Latent Sketchpad~\citep{zhang2025latentsketchpadsketchingvisual} exemplify this by empowering models to generate and refine mental imagery during the inference process, effectively bridging the gap between visual grounding and abstract logic. Further advancements include Latent Visual Reasoning (LVR)~\citep{li2025latentvisualreasoning} and implicit reasoning tokens~\citep{li2025latentimplicitvisualreasoning}, which allow models to perform autoregressive planning in the visual embedding space without explicit supervision. To effectively untangle complex visual environments, approaches like ROSS~\citep{wang2024reconstructive} complement this paradigm by explicitly modeling spatial topological structures within the latent space via a denoising objective to reconstruct latent representations of the question image as vision-centric supervision, augmenting the model's structural comprehension. These methods overcome the limitations of text-centric reasoning by maintaining rich, high-dimensional visual information throughout the decision-making chain~\citep{li2025latentvisualreasoning, li2025latentimplicitvisualreasoning}. Furthermore, similar principles of latent semantic grounding have also driven advances in continuous 3D spatial modeling~\citep{yu2025core3d,yu2026elsa3d,yu2026dreampartgen, yu2026silsa} and generative modeling~\citep{fang2025signllm, fang2025signdiff, fang2026stable, fang2025signx, fang2026signverse, fang2025streamflow, fang2026rac, li2026echofoley}. While these prior works demonstrate the potential of utilizing latent spaces for intermediate reasoning, they mainly rely on either unconstrained autoregressive planning or vision-centric supervision derived from the input context. In contrast, our work predicts visual hypotheses conditioned on intermediate reasoning prefixes, and the latent-rationale optimization stage uses their consistency to shape the reasoning policy.

\section{Theoretical Grounding of the Five Cognitive Dimensions}
\label{appx_sec:grounding}
\setcounter{figure}{0}
\setcounter{table}{0}

\begin{table*}[!t]
    \centering
    \caption{\textbf{CogSense Benchmark Dimensions and Psychometric Constructs.}
    We define five CogSense dimensions based on established psychometric constructs and cognitive theories, with each dimension comprising multiple cognitive subsets.}

    \label{tab:cogsense_dimensions}
    \setlength{\tabcolsep}{6pt}
    \renewcommand\arraystretch{1.15}
    \resizebox{\linewidth}{!}{%
      \begin{tabular}{l c l l}
        \toprule
        \rowcolor{gray!25}
        \textbf{Dimension} &
        \textbf{Psychometric Construct} &
        \textbf{Source} &
        \textbf{Subsets} \\
        \midrule

        Fluid Intelligence
        & $G_\mathrm f$
        & \makecell[l]{Theory of Fluid and Crystallized Intelligence~\citep{cattell1963theory}, \\
        Structure-mapping~\citep{gentner1983structure}}
        & \makecell[l]{MaRs-VQA, \\ PGM, RAVEN} \\
        \midrule

        Crystallized Intelligence
        & $G_\mathrm c$
        & \makecell[l]{Theory of Fluid and Crystallized Intelligence~\citep{cattell1963theory}, \\
        Natural Categories~\citep{rosch1973natural}}
        & \makecell[l]{Bongard-RWR+, \\ Bongard-HOI} \\
        \midrule

        Visuospatial Cognition
        & $G_\mathrm v$
        & \makecell[l]{Untersuchungen zur Lehre von der Gestalt~\citep{wertheimer1923untersuchungen}, \\
        Recognition-by-components~\citep{Biederman1987Recognition}}
        & Bongard-LOGO \\
        \midrule

        Mental Simulation
        & \makecell[l]{$G_\mathrm v/G_\mathrm f\ \cap$ \\ Causal Reasoning}
        & Simulation as an Engine of Physical Scene Understanding~\citep{Battaglia2013simulation}
        & \makecell[l]{KiVA, STARE, \\ ARC-AGI, ARC-AGI-2} \\
        \midrule

        Visual Routines
        & \makecell[l]{$G_\mathrm v/G_\mathrm f\ \cap$ \\ Sequential Attention}
        & Visual Routines~\citep{ullman1984visual}
        & CVR \\

        \bottomrule
      \end{tabular}
    }
\end{table*}

As shown in~\Cref{tab:cogsense_dimensions}, the first three dimensions represent the traditional psychometric approach, while the last two dimensions belong to the process-level tradition of visual cognition. Together, these dimensions cover complementary operations required for high-level visual cognition: inducing rules, retrieving concepts, organizing spatial structure, simulating changes, and executing attention-based visual operations. We exclude other traditional psychometric dimensions such as Working Memory ($G_\mathrm{wm}$) and Processing Speed ($G_\mathrm s$). These dimensions measure cognitive capacity and execution speed rather than reasoning operations. In the context of MLLMs, they are largely confounded with architectural hyperparameters (\eg context window length and KV-cache) rather than inner visual cognitive capabilities.\looseness-1

\section{Statistics of CogSense-Bench}
\label{appx_sec:data_stats}
\setcounter{figure}{0}
\setcounter{table}{0}

\begin{table}[!t]
    \centering
    \caption{\textbf{CogSense-Bench descriptions and statistics.} CogSense-Bench comprises multiple datasets grouped into five cognitive categories, covering diverse aspects of visual reasoning, including fluid intelligence, crystallized intelligence, visuospatial cognition, mental simulation, and visual routines.  We report the number of training and test VQA pairs. Dataset licenses are provided when available.}
    \label{tab:appx_dataset_stats}

    \resizebox{0.8\linewidth}{!}{%
        \begin{tabular}{llccl}
            \toprule
            \rowcolor{gray!25}
            \textbf{Category} & \textbf{Dataset} & \textbf{\# Train} & \textbf{\# Test} & \textbf{License} \\
            \midrule
            \multirow{3}{*}{Fluid Intelligence}
            & MaRs-VQA & 1.4K & 13 & CC-BY-NC-3.0 \\
            & PGM & 10K & 94 & Unspecified \\
            & RAVEN & 18K & 169 & GPL-3.0 License \\
            \midrule
            \multirow{2}{*}{Crystallized Intelligence}
            & Bongard-RWR+ & 16K & 152 & CC-BY-4.0 \\
            & Bongard-HOI & 23K & 216 & NVIDIA Source Code License \\
            \midrule
            \multirow{1}{*}{Visuospatial Cognition}
            & Bongard-LOGO & 12K & 113 & MIT license \\
            \midrule
            \multirow{4}{*}{Mental Simulation}
            & KiVA & 1.4K & 13 & Apache-2.0 License \\
            & STARE & 4K & 26 & Unspecified \\
            & ARC-AGI & 1.6K & 31 & Apache-2.0 License \\
            & ARC-AGI-2 & 8K & 80 & Apache-2.0 License \\
            \midrule
            \multirow{1}{*}{Visual Routines}
            & CVR & 10K & 93 & Apache-2.0 License \\
            \midrule
            \multirow{1}{*}{\textbf{Sum}}
            &  & 105.4K & 1000 &  \\
            \bottomrule
        \end{tabular}
    }
\end{table}

We summarize the statistics of CogSense-Bench in~\cref{tab:appx_dataset_stats}. Throughout the paper, Avg. denotes the overall accuracy over all 1,000 test items, \ie, the category accuracies weighted by the number of test items in each category, rather than the unweighted mean of the five categories. We ensured zero data leakage by randomly sampling instances from each category at the same proportion as in the train split and subsequently removing the sampled data from it. \looseness=-1

\section{Implementation Details for Data Pipeline}
\label{appx_sec:data_pipeline}
\setcounter{figure}{0}
\setcounter{table}{0}

\subsection{Data Extraction}
We curate visual cognition data from a broad set of datasets spanning Bongard-style reasoning, abstract visual reasoning, spatial cognition, and ARC-style program induction, including Bongard-OpenWorld~\citep{wu2024bongardopenworld}, Bongard-HOI~\citep{jiang2022bongard}, Bongard-LOGO~\citep{nie2020bongard}, Bongard-RWR+~\citep{pawlonka2025bongardrwrrealworldrepresentationsfinegrained}, CVR~\citep{zerroug2022benchmark}, KiVA~\citep{yiu2025kiva}, MaRs-VQA~\citep{cao2025visualcognitiongaphumans}, PGM~\citep{barrett2018measuring}, RAVEN~\citep{zhang2019raven}, I-RAVEN~\citep{hu2022stratifiedruleawarenetworkabstract}, RAVEN-FAIR~\citep{benny2021scale}, A-I-RAVEN~\citep{malkinski2025airaven}, I-RAVEN-Mesh~\citep{malkinski2025airaven}, I-RAVEN-X~\citep{camposampiero2025iravenxbenchmarkinggeneralizationrobustness}, STARE~\citep{li2025unfoldingspatialcognitionevaluating}, ARC-AGI~\citep{chollet2019measure}, and ARC-AGI-2~\citep{chollet2025arcagi2newchallengefrontier}. From these sources, we manually select examples that match our five predefined cognitive categories, ensuring that each retained instance is relevant to the target reasoning skills evaluated in our benchmark.
Building on these seed sources, we develop automated extraction and cleaning scripts to expand the collection at scale and produce the raw dataset. To standardize evaluation across heterogeneous task formats, we further convert tasks that were not originally multiple-choice into a unified multiple-choice format. For Bongard-style problems, we randomly select one image from the positive set as the correct answer, combine it with negative samples, and shuffle the resulting candidates as answer options; the remaining positive samples are used to define the query context. For ARC-AGI-style problems, we perform data augmentation on the ground-truth images and present both the original ground-truth image and augmented variants as candidate options.

\begin{figure}[!p]
\centering
\resizebox{0.97\textwidth}{!}{
\begin{planbox}{Reasoning Chain Generation Prompts for CogSense-Bench}

\

\small

You are a careful visual reasoning assistant. Your task is to generate CoT and the final answer for the provided questions.

Your response must consist of \textbf{ONLY} the following two parts in JSON format:
\begin{enumerate}
    \item `reasoning': reasoning steps (3-12 concise steps from high-level to details).
    \item `final\_answer': the final answer of the question. Keep reasoning concise (3-12 steps). \textbf{Do NOT} include internal deliberation beyond the steps.
\end{enumerate}

Output must be in JSON format and \textbf{ONLY} with the reasoning steps and final answer with nothing else.

\hrulefill

\emph{For Crystallized Intelligence and Visuospatial Cognition Questions:}

You will be given a set of images that share a common pattern. Analyze these examples to understand the pattern. Then, select the image that belongs to the same category as the images in the question set.

Question images (these images share a common pattern): [question\_images]

Option images: [option\_images]

\hrulefill

\emph{For Fluid Intelligence Questions:}

You will be given an image that contains eight subimages and a blank space. Then, given some option images, predict which one is the right option, following the same pattern to complete the sequence based on the pattern shown in the question image.

Question image: <question\_image>

Option images: [option\_images]

\hrulefill

\emph{For ARC-AGI in Mental Simulation:}

You will be given some pairs of input-output images as examples. The output is obtained by coloring the grid according to a specific pattern based on the input. Analyze these examples to understand the underlying grid color pattern. Then, given a new input image and option images, predict which image is generated by applying the same pattern to the input.

Examples: [<input\_image, output\_image>]

Question image: <new\_input\_image>

Option images: [option\_images]

\hrulefill

\emph{For KiVA in Mental Simulation:}

You will be given some pairs of input-output images or transform patterns as examples. Analyze the pattern and select the correct answer from the given options. Then, given a new input image and some options that indicate possible outputs, predict the right option following the same pattern.

Examples: [pattern\_image]

Question image: <new\_input\_image>

Option images: [option\_images]

\hrulefill

\emph{For STARE in Mental Simulation:}

You will be given some images and a blank space. Then, given some option images, predict which one is the right option, following the same pattern to complete the sequence based on the pattern shown in the
question image.

Question image: <question\_image>

Option images: [option\_images]

\hrulefill

\emph{For Visual Routines Questions:}

You will be given some images, in which one of them differs from the rest of the images. Find the odd one out.

Question images: [images]

Option images: [images]

\hrulefill

\textbf{Input:} [images] and <prompt>

\textbf{Output:} <reasoning\_chain> and <answer>

\end{planbox}
}

\caption{\textbf{Reasoning Chain Generation Prompts.} We customize different prompts for different types of original questions and use LLMs to generate corresponding reasoning chains based on these different prompts.}
\label{fig:appx_prompt}
\end{figure}

\subsection{Reasoning Chain Generation}
\label{appx_data_pipeline_subsec:reasoning}
We transform traditional short QA pairs into high-quality reasoning data containing an explicit reasoning chain. We customize different prompts for different types of original questions and leverage LLMs to generate corresponding reasoning chains based on these different prompts. The teacher model $\mathcal{M}_T$ is required not only to output the final answer but also to elaborate on its reasoning process. Specifically, it is guided to analyze visual cues within the images, infer underlying rules, and explain causal relationships, finally generating detailed reasoning process text. To ensure quality, we employ an automated verification pipeline: (1) The teacher model generates the rationale and a final prediction based on task-specific prompt protocols (\Cref{fig:appx_prompt}); (2) We automatically parse the final predicted answer and match it strictly against the ground truth; (3) If the prediction is incorrect, the reasoning trace is immediately discarded, and we prompt the teacher model to retry (up to $K = 3$ times); (4) For the subset of highly complex questions where the teacher model consistently failed even after 3 retries, we manually solved them and explicitly authored the step-by-step reasoning traces; (5) For the teacher-generated subset that successfully passed the automated answer-consistency filter, we conducted a human evaluation to ensure the generated rationales do not contain hallucinations. We randomly sampled a subset of 1,000 verified reasoning traces for a blind review by human annotators. The evaluation showed an overall acceptance rate of \textbf{95.4\%}, demonstrating that our automated answer-consistency filtering highly correlates with rationale quality.

\section{Human Study Design and Setup}
\label{appx_sec:human_study}

\paragraph{Participant Recruitment} To establish a human performance baseline, we recruited a total of 20 participants through online platforms using a random sampling strategy, with ages ranging from 18 to 34 (average 24.5). They participated purely out of interest in AI cognitive evaluation and to test their own performance against the benchmark. Participation in the study was entirely voluntary, and no financial compensation or monetary incentives were provided to the subjects.

\paragraph{Questionnaire Design} The questionnaire design involved a rigorous selection process to ensure representativeness. We constructed a test set comprising 100 multiple-choice questions derived from the CogSense-Bench. These questions were manually selected by stratified sampling to strictly maintain the category proportions of the original benchmark. The study was implemented and distributed digitally via the Google Forms platform, allowing for remote administration.

\paragraph{Time Spent and Compensation} The survey took participants an average of 45 minutes to complete. Participants proceeded entirely voluntarily and were free to drop out at any point without consequence. To ensure data quality, we monitored response times and confirmed that none of the participants rushed through the survey. The high consistency in their answers also indicates rigorous engagement with the task. Since uncompensated volunteer recruitment limits the representativeness of the human baseline, we acknowledge this limitation and do not present the 20-participant result as a population-level estimate of human cognitive performance.

\paragraph{Consent Procedure} Before evaluation, all participants were presented with a digital informed consent form that outlined the purpose of the study, the estimated time commitment, the voluntary and uncompensated nature of the task, and the data anonymity protocols. Participants were required to explicitly agree to these terms before proceeding. The study protocol, including the consent process and volunteer framework, was reviewed and approved by our Institutional Review Board (IRB) under an existing IRB approval for visual cognition research at our institution.

\paragraph{Compliance} Regarding experimental standards and ethical compliance, all protocols were reviewed and approved by the IRB prior to the study.

\section{Comparison between ROSS and CogSense}
\label{appx_sec:comparision}
\setcounter{figure}{0}
\setcounter{table}{0}

\begin{table*}[!t]
  \centering
  \caption{\textbf{Performance Comparison under Cambrian.} All models are based on the Qwen3-VL-8B backbone and trained exclusively on the Cambrian-737K dataset.}
  \label{tab:compare1}

  \setlength{\tabcolsep}{5.3pt}
  \renewcommand\arraystretch{1.15}

  \rowcolors{2}{gray!10}{white}

  \footnotesize
  \resizebox{\linewidth}{!}{%
  \begin{tabular}{lccccc}
    \toprule
    \rowcolor{gray!25}
    \textbf{Model} & \textbf{HallusionBench} & \textbf{MMBench-EN} & \textbf{MMBench-CN} & \textbf{MMMU} & \textbf{AI2D} \\
    \midrule
    ROSS (Qwen3-VL-8B, Cambrian) & 59.6 & 80.6 & 81.2 & 64.3 & 79.9 \\
    CogSense (Qwen3-VL-8B, Cambrian) \textit{w/o} LVIP & 60.7 & 81.5 & 81.7 & 69.6 & 82.7 \\
    CogSense (Qwen3-VL-8B, Cambrian) & \textbf{62.5} & \textbf{83.7} & \textbf{84.1} & \textbf{71.1} & \textbf{84.2} \\
    \bottomrule
  \end{tabular}}
\end{table*}

We compare CogSense against ROSS~\citep{wang2024reconstructive} by conducting two more experiments, where we train the same base model Qwen3-VL-8B~\citep{bai2025qwen3vltechnicalreport} with both frameworks. For ROSS, we use the same denoiser as~\citep{wang2024reconstructive}, freeze Qwen3-VL-8B, and pre-train the denoiser with the same teacher tokenizer under a mixture of 2M caption data (1246K from ShareGPT4V~\citep{chen2024sharegpt4v} and 707K from ALLaVA~\citep{chen2024allava}). For CogSense, we do not perform any pre-training.

First, we train both models with supervised fine-tuning on Cambrian-737K~\citep{tong2024cambrian}. For CogSense, we further train the model with the proposed RL method on the same dataset. We evaluate both models on the listed benchmarks. We further ablate our method with and without LVIP under this setting. We then evaluate this newly trained model across a range of established general-domain and fine-grained multimodal benchmarks like HallusionBench~\citep{guan2024hallusionbenchadvancediagnosticsuite}, MMBench~\citep{liu2024mmbench}, MMMU~\citep{yue2024mmmu}, and AI2D~\citep{kembhavi2016diagramworthdozenimages}. In addition, we train Qwen3-VL-8B on ROSS settings with supervised fine-tuning on the train split of CogSense-Bench and then evaluate both models on the test split.

As shown in~\Cref{tab:compare1} and~\Cref{tab:compare2}, first, under the same Qwen3-VL-8B backbone and Cambrian settings, adding LVIP consistently improves performance across all five general-domain benchmarks. Moreover, under the same setting, Qwen3-VL-8B backbone, and dataset, our framework also outperforms ROSS. These results isolate the effect of LVIP from backbone or data differences, showing that LVIP provides robust gains beyond our cognitive benchmark and transfers to broader multimodal understanding and fine-grained visual reasoning.

\begin{table*}[!t]
  \centering
  \caption{\textbf{Performance Comparison under CogSense-Bench.} All models are based on the Qwen3-VL-8B backbone and trained exclusively on the train split of CogSense-Bench.}
  \label{tab:compare2}

  \setlength{\tabcolsep}{5.3pt}
  \renewcommand\arraystretch{1.15}

  \rowcolors{2}{gray!10}{white}

  \footnotesize
  \resizebox{\linewidth}{!}{%
  \begin{tabular}{lcccccc}
    \toprule
    \rowcolor{gray!25}
    \textbf{Model} & \textbf{Fluid Intel.} & \textbf{Crystallized Intel.} & \textbf{Visuospatial Cog.} & \textbf{Mental Simu.} & \textbf{Visual Rout.} & \textbf{Avg.} \\
    \midrule
    ROSS (Qwen3-VL-8B) & 54.0 & 84.8 & 63.7 & 60.7 & 43.0 & 66.4 \\
    \ourmodel (Qwen3-VL-8B) & \textbf{63.8} & \textbf{91.0} & \textbf{69.0} & \textbf{68.0} & \textbf{50.5} & \textbf{73.8} \\
    \bottomrule
  \end{tabular}}
\end{table*}

\section{Statistical Significance Analysis}
\label{appx_sec:analysis}
\setcounter{figure}{0}
\setcounter{table}{0}

\paragraph{Bootstrap Confidence Intervals} We compute 95\% confidence intervals using a stratified paired bootstrap over the benchmark. Each bootstrap sample preserves the category proportions of CogSense-Bench. We use 10,000 bootstrap resamples.

\begin{table*}[!t]
    \centering
    \caption{\textbf{Main Results with 95\% CI on CogSense-Bench.} Confidence
    Intervals are computed by a stratified paired bootstrap over benchmark items.}
    \label{tab:CI}
      \setlength{\tabcolsep}{5.3pt}
    \renewcommand\arraystretch{1.15}
    \rowcolors{2}{gray!10}{white}
    \footnotesize
    \resizebox{0.60\linewidth}{!}{%
        \begin{tabular}{lcc}
            \toprule
            \rowcolor{gray!25}
            \textbf{Model} & \textbf{Accuracy} & \textbf{95\% CI} \\
            \midrule
            \textcolor{humangray}{\textbf{Human}} & 88.4 & [85.5, 91.3] \\
            \midrule
            Gemini 2.5 Flash & 34.9 & [32.0, 37.8] \\
            GPT-o3 & 32.3 & [29.4, 35.2] \\
            GPT-5.2 & 40.3 & [37.4, 43.2] \\
            Claude Sonnet 4 & 32.6 & [29.7, 35.5] \\
            Grok 4 Fast & 31.7 & [28.8, 34.6] \\
            \midrule
            Llama-4-Scout-17B & 31.8 & [28.9, 34.7] \\
            Gemma-3-27B & 32.7 & [29.8, 35.6] \\
            Ministral-3-14B & 16.5 & [13.6, 19.4] \\
            Intern-S1 & 17.5 & [14.6, 20.4] \\
            Qwen3-VL-30B & 37.4 & [34.5, 40.3] \\
            \midrule
            InternVL3.5-8B (FT) & 63.9 & [61.0, 66.8] \\
            Qwen3.5-9B (FT) & 69.1 & [66.2, 72.0] \\
            \midrule
            Qwen3-VL-8B (Base) & 35.5 & [32.6, 38.5] \\
            Qwen3-VL-8B SFT \textit{w/o} LVIP & 62.3 & [59.3, 65.3] \\
            Qwen3-VL-8B SFT \textit{w/} LVIP & 68.0 & [65.1, 70.8] \\
            Qwen3-VL-8B SFT \textit{w/o} LVIP + GRPO & 65.5 & [62.6, 68.4] \\
            Qwen3-VL-8B SFT \textit{w/} LVIP + GRPO & 70.8 & [68.0, 73.6] \\
            Qwen3-VL-8B SFT \textit{w/} LVIP + GRPO + LR & 71.5 & [68.6, 74.2] \\
            \midrule
            \ourmodel (Ours) & 73.8 & [71.0, 76.4] \\
            \bottomrule
        \end{tabular}
    }
\end{table*}

\paragraph{McNemar Test} For pairwise comparison between \ourmodel and a baseline, we report McNemar's test~\citep{mcnemar1947note} on paired correctness outcomes over all $N$ benchmark instances. Let $b$ denote the number of instances that \ourmodel gets correct while the baseline gets wrong, and $c$ denote the number of instances that \ourmodel gets wrong while the baseline gets correct. The test is conducted on the $n_{\mathrm{d}}=b+c$ discordant pairs. We compute the two-sided exact McNemar $p$-value using a binomial test under the null hypothesis that both models have equal accuracy, i.e., $b \sim \mathrm{Binomial}(n_{\mathrm{d}}, 0.5)$. The $p$-value is defined as the probability of observing a value at least as extreme as the observed $b$ under this distribution.

Our statistical analysis validates the efficacy of our specific training pipeline and architectural choices. As detailed in the paired performance comparisons, \ourmodel significantly outperforms the strongest same-backbone baseline (Qwen3-VL-8B SFT \textit{w/} LVIP + GRPO + LR, $\Delta = +2.3\%$) under both the stratified paired bootstrap (95\% CI: [0.7, 3.9]) and the exact McNemar test ($p = 0.0265$). This supports that the observed improvement is unlikely to be explained by benchmark sampling noise alone, confirming the robustness and effectiveness of our proposed approach.

\begin{table*}[!t]
    \centering
    \caption{\textbf{Paired Performance Comparison Results.} We report bootstrap confidence intervals and McNemar's test results between \ourmodel and strong baselines.}

    \label{tab:pair}
    \setlength{\tabcolsep}{5.3pt}
    \renewcommand\arraystretch{1.15}
    \rowcolors{2}{gray!10}{white}
    \footnotesize
    \resizebox{0.9\linewidth}{!}{%
        \begin{tabular}{lcccc}
            \toprule
            \rowcolor{gray!25} \textbf{Comparison (Ours vs. ...)} & \textbf{$\Delta$ Acc.} & \textbf{Bootstrap 95\% CI} & \textbf{$(b,c)$} & \textbf{McNemar $p$} \\
            \midrule
            InternVL3.5-8B (FT) & +9.9 & [7.4, 12.4] & (145, 46) & $<10^{-4}$ \\
            Qwen3.5-9B (FT) & +4.7 & [2.6, 6.8] & (100, 53) & 0.0002 \\
            \midrule
            Qwen3-VL-8B (Base) & +38.3 & [35.2, 41.4] & (420, 37) & $<10^{-4}$ \\
            Qwen3-VL-8B SFT \textit{w/o} LVIP & +11.5 & [8.9, 14.1] & (160, 45) & $<10^{-4}$ \\
            Qwen3-VL-8B SFT \textit{w/} LVIP & +5.8 & [3.7, 8.1] & (121, 63) & $<10^{-4}$ \\
            Qwen3-VL-8B SFT \textit{w/o} LVIP + GRPO & +8.3 & [6.0, 10.6] & (130, 47) & $<10^{-4}$ \\
            Qwen3-VL-8B SFT \textit{w/} LVIP + GRPO & +3.0 & [1.2, 4.8] & (78, 48) & 0.0095 \\
            Qwen3-VL-8B SFT \textit{w/} LVIP + GRPO + LR & +2.3 & [0.7, 3.9] & (61, 38) & 0.0265 \\
            \bottomrule
        \end{tabular}
    }
\end{table*}

\section{Additional Qualitative Examples}
\label{appx_sec:qualitative_examples}
\setcounter{figure}{0}
\setcounter{table}{0}

\begin{figure*}[t]
    \centering
    \includegraphics[width=0.99\linewidth]{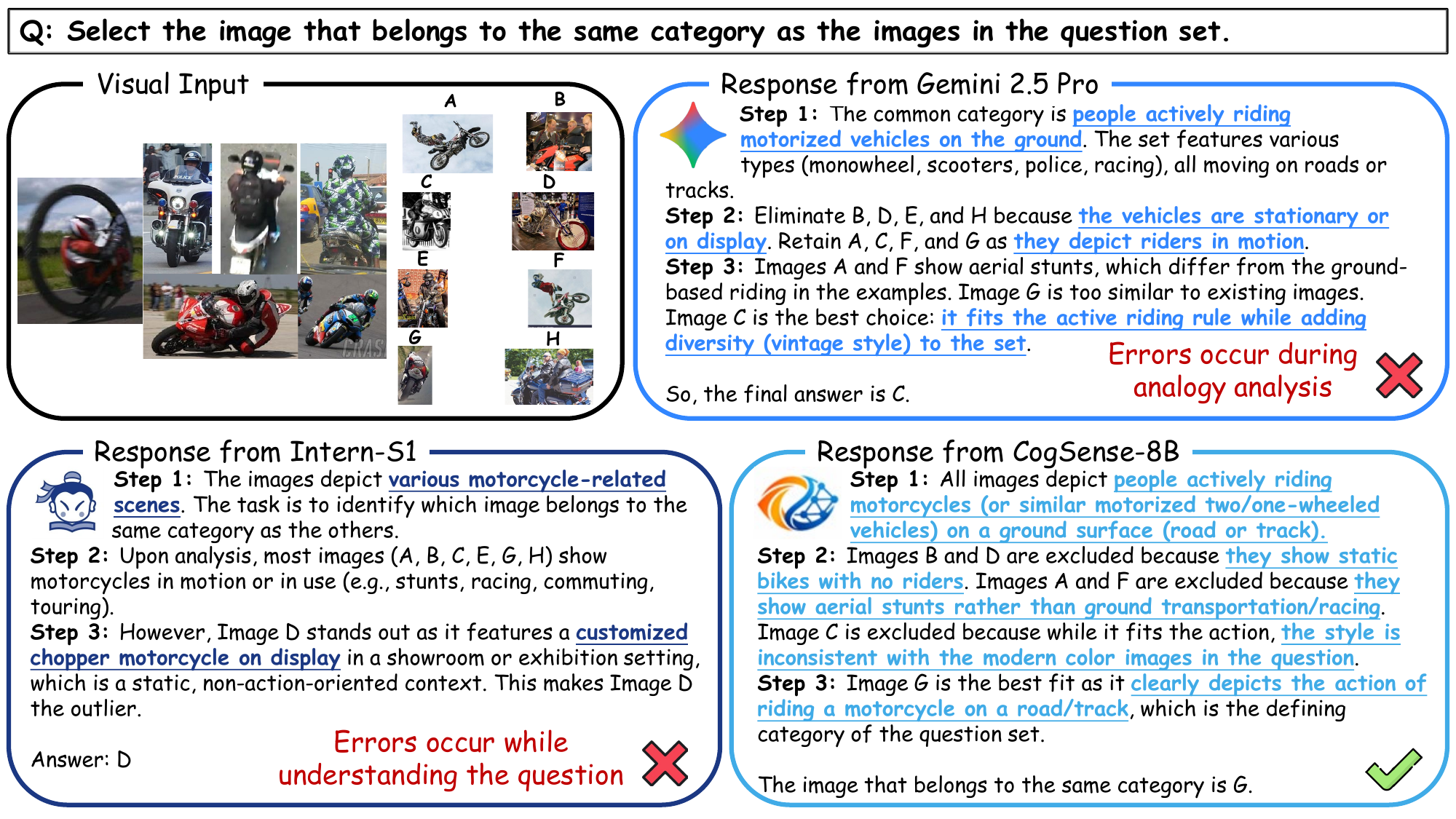}
    \caption{\textbf{Additional Qualitative Examples of Visual Cognition Reasoning Across Models (1/2).} \ourmodel demonstrates a coherent, multi-step logical chain that closely matches the ground truth, while other models exhibit less precise or less interpretable reasoning paths.}
    \label{fig:appx_qualitative_example_1}
\end{figure*}

\begin{figure*}[t]
    \centering
    \includegraphics[width=0.99\linewidth]{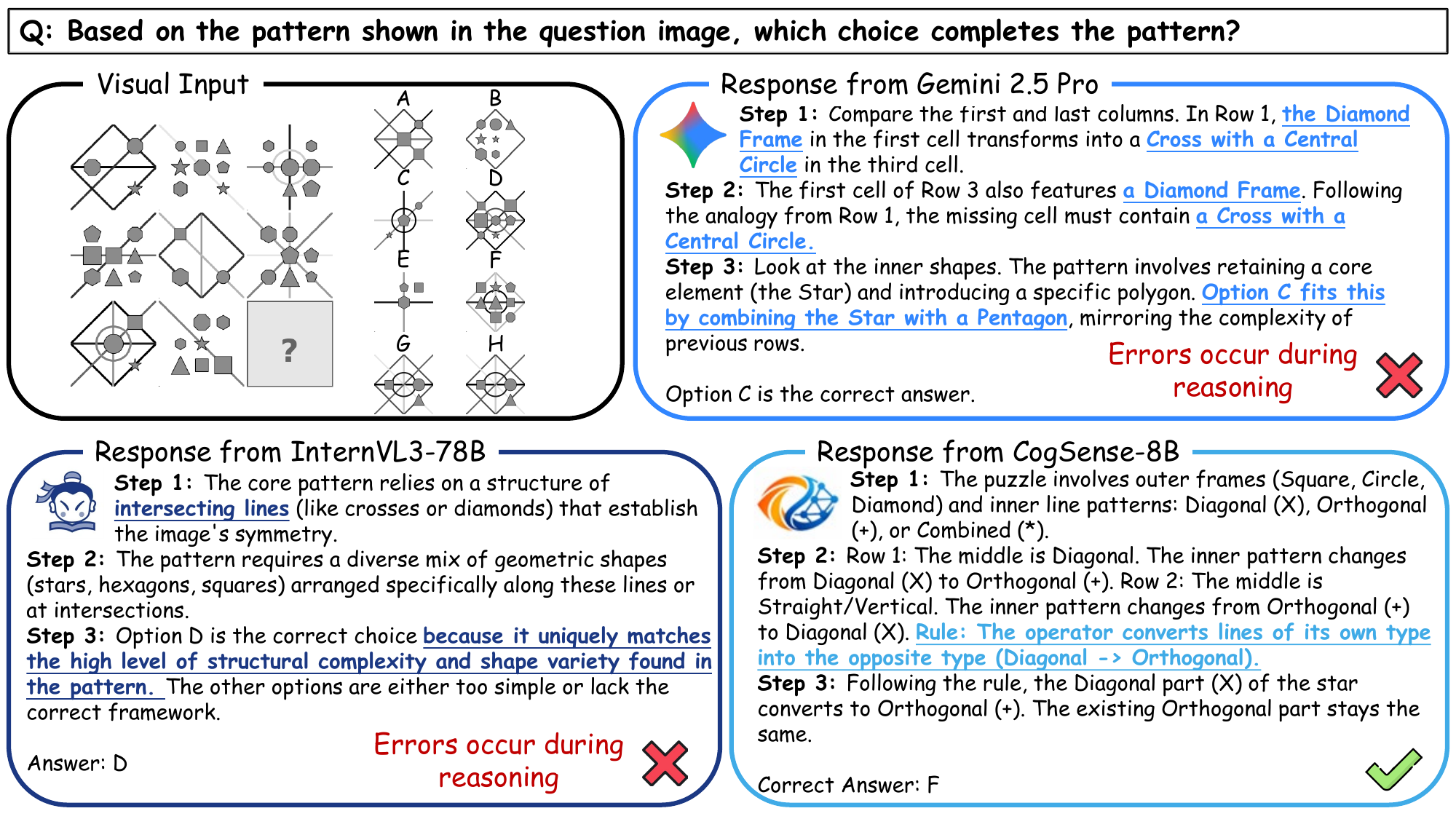}
    \caption{\textbf{Additional Qualitative Examples of Visual Cognition Reasoning Across Models (2/2).} \ourmodel demonstrates a coherent, multi-step logical chain that closely matches the ground truth, while other models exhibit less precise or less interpretable reasoning paths.}
    \label{fig:appx_qualitative_example_2}
\end{figure*}

More qualitative examples are shown in~\cref{fig:appx_qualitative_example_1,fig:appx_qualitative_example_2}. These examples demonstrate that \ourmodel exhibits a superior capacity for high-level visual reasoning and rule induction compared to both large-scale proprietary models like Gemini 2.5 Pro and specialized baselines such as InternVL3. In complex categorization tasks, while baseline models often struggle with task comprehension or fail to distinguish between superficial visual similarities and core semantic categories, \ourmodel accurately identifies latent action-based rules through analysis.

In abstract geometric reasoning, \ourmodel also successfully deciphers complex transformations between orthogonal and diagonal line structures, where other models fail to move beyond basic structural complexity. These results highlight the model's robustness in bridging the gap between low-level perception and multi-step cognitive reasoning, particularly in scenarios requiring precise spatial logic.

\section{Additional General Ability Evaluation}
\label{appx_sec:general}
\setcounter{figure}{0}
\setcounter{table}{0}

\begin{table*}[t!]
  \centering
  \caption{\textbf{More General Ability Results.} Performance comparison in general ability in vision-language understanding and spatial reasoning benchmarks with \ourmodel and the base model. \ourmodel maintains robust performance on these general benchmarks, comparable to the base model.}
  \label{tab:exp_more_general}

  \setlength{\tabcolsep}{5.3pt}
  \renewcommand\arraystretch{1.15}

  \rowcolors{2}{gray!10}{white}

  \footnotesize
  \resizebox{0.99\linewidth}{!}{%
  \begin{tabular}{lccccccccccc}
    \toprule
    \rowcolor{gray!25}
    \textbf{Model} & \textbf{AI2D} & \textbf{ChartQA} & \textbf{MMStar} & \textbf{MMMU} & \textbf{MMMU-Pro} & \textbf{VisualPuzzles} & \textbf{MindCube} & \textbf{SITE} & \textbf{MathVista} & \textbf{MathVerse} \\
    \midrule
    Qwen3-VL-8B (Base) & \textbf{85.4} & \textbf{88.6} & \textbf{70.9} & 69.6 & 55.9 & 47.9 & 41.5 & 41.1 & 77.2 & 62.1 \\
    \ourmodel (Ours) & 85.1 & 84.7 & 66.8 & \textbf{71.4} & \textbf{58.2} & \textbf{53.6} & \textbf{42.5} & \textbf{42.1} & \textbf{79.1} & \textbf{65.7} \\
    \bottomrule
  \end{tabular}}
\end{table*}

We further compare \ourmodel against the base model on: (1) general vision-language benchmarks, including AI2D~\citep{kembhavi2016diagramworthdozenimages}, ChartQA~\citep{masry2022chartqabenchmarkquestionanswering}, MMStar~\citep{chen2024rightwayevaluatinglarge}, MMMU~\citep{yue2024mmmu}, MMMU-Pro~\citep{yue2025mmmu}, and VisualPuzzles~\citep{song2025visualpuzzles}; (2) spatial reasoning benchmarks, including MindCube~\citep{wang2025mindcube} and SITE~\citep{wang2025site}; and (3) mathematical reasoning benchmarks, including MathVista~\citep{lu2024mathvista} and MathVerse~\citep{zhang2024mathverse}. As shown in~\Cref{tab:exp_more_general}, \ourmodel remains competitive with the base model on these general-domain benchmarks, indicating that it does not overfit to the cognitive training distribution.
As is commonly observed in post-training, performance fluctuations can arise across diverse downstream tasks (some may increase and some may decrease). In particular, ChartQA and MMStar rely heavily on OCR-related capabilities of VLMs and require fine-grained text extraction capabilities. This differs from our target domain of pure visual reasoning, which does not rely on textual priors embedded within images. More importantly, \ourmodel outperforms the same-scale base model on all newly introduced spatial benchmarks, demonstrating stronger spatial generalization and more robust reasoning across evaluation settings. These results show that the architectural and training components introduced in \ourmethod yield consistent gains over the base model in the target reasoning domains while preserving broad multimodal competence.

\section{Discussion on LVIP Mechanism and Latent Visual Imagery Learning}
\label{appx_sec:lvip}
\setcounter{figure}{0}
\setcounter{table}{0}

\paragraph{LVIP Mechanism Explanation} The latent visual imagery generation is synchronized with the autoregressive next-token prediction. At each decoding step of the textual CoT, the LVIP head generates a corresponding visual latent embedding based on the current context. Therefore, the LVIP head produces a sequence of latent visual features that changes as the reasoning progresses. The generated latent visual imagery does not merely passively follow the text's line of reasoning but actively corrects it. As detailed in~\Cref{sec:method}, LVIP is evaluated at intermediate prefixes $[X; z_{1:t}]$, and prefixes whose latent visual states become inconsistent receive lower reward $R_{\mathrm{lvip}}(\tau_t; X, y)$ and therefore lower probability under the learned policy $q_{\theta}(Z \mid X)$. In this way, LVIP influences reasoning by guiding the model toward trajectories that remain visually consistent. Through the Sub-Trajectory Balance (SubTB)~\citep{madan2023learninggflownetspartialepisodes} loss, the model is explicitly trained to assign a lower probability flow to these branching paths. In this way, LVIP trains the policy to avoid those wrong reasoning paths, and also prevents the answer-conditioned shortcut behavior through prefix-level optimization.

\paragraph{Analysis of Latent Visual Imagery Evolution} We track the trajectories of the intermediate latent visual imagery states during the reasoning process when the final answer is correct/incorrect. Specifically, we measure the MSE loss between the intermediate latent visual imagery $\hat{\mathbf{h}}_{y, i}$ at various steps of the CoT and the final ground-truth latent visual imagery $\mathbf{h}_y$. The results are averaged over 10 sampled trajectories. As shown in~\Cref{tab:exp_track}, for correct answer trajectories, the MSE loss decreases as the step increases, providing evidence that the latent visual imagery changes and converges toward the correct GT step-by-step alongside the reasoning. For incorrect answer trajectories, the MSE fails to converge to the ground-truth latent state. When the textual reasoning chain begins to hallucinate or drift into the wrong logical path, the latent visual features also reflect this internal misunderstanding.

\paragraph{Ablation Study on Chain-of-Thought} To validate that latent visual imagery relies on the step-by-step textual reasoning, we conduct an ablation study removing the CoT, forcing the model to generate the final answer directly in a single-step reply. As demonstrated in~\Cref{tab:exp_cot}, removing the CoT leads to a performance drop, demonstrating that the intermediate textual reasoning is crucial to guide and refine the latent visual imagery toward the correct answer.

\begin{table}[!t]
    \centering
    \caption{\textbf{Trajectories of Intermediate Latent Visual States.} We report the MSE loss between intermediate and ground-truth latent imagery across different reasoning steps. For correct answers, the latent state progressively converges, whereas it fails to converge for incorrect answers, demonstrating dynamic synchronization alongside textual reasoning.}
    \label{tab:exp_track}

    \setlength{\tabcolsep}{6pt}
    \renewcommand\arraystretch{1.15}
    \rowcolors{2}{gray!10}{white}
    \resizebox{0.6\linewidth}{!}{%
      \begin{tabular}{lccccc}
        \toprule
        \rowcolor{gray!25}
        \textbf{Final Answer} & \textbf{Step 0} & \textbf{Step 10} & \textbf{Step 20} & \textbf{Step 30} & \textbf{Step 40} \\
        \midrule
        Correct Answer & 0.492 & 0.368 & 0.215 & 0.184 & 0.158 \\
        Incorrect Answer & 0.495 & 0.442 & 0.405 & 0.438 & 0.426 \\
        \bottomrule
      \end{tabular}
    }
\end{table}

\begin{table*}[!t]
    \centering
    \caption{\textbf{CoT Ablation Study Results.} We compare the performance of \ourmodel \textit{w/} and \textit{w/o} intermediate CoT across CogSense-Bench. Forcing the model to generate the final answer directly in a single step (\textit{w/o} CoT) leads to a consistent performance drop, demonstrating that step-by-step textual reasoning is crucial for guiding the latent visual imagery.}
    \label{tab:exp_cot}

    \setlength{\tabcolsep}{6pt}
    \renewcommand\arraystretch{1.15}
    \rowcolors{2}{gray!10}{white}
    \resizebox{\linewidth}{!}{%
      \begin{tabular}{lcccccc}
        \toprule
        \rowcolor{gray!25}
        \textbf{Variant} & \textbf{Fluid Intel.} & \textbf{Crystallized Intel.} & \textbf{Visuospatial Cog.} & \textbf{Mental Simu.} & \textbf{Visual Rout.} & \textbf{Avg.} \\
        \midrule
        \ourmodel \textit{w/o} CoT & 61.6 & 87.0 & 66.4 & 65.3 & 45.2 & 70.5 \\
        \ourmodel \textit{w/} CoT & 63.8 & 91.0 & 69.0 & 68.0 & 50.5 & 73.8 \\
        \bottomrule
      \end{tabular}
    }
\end{table*}

\begin{table*}[!t]
    \centering
    \caption{\textbf{Prefix Sensitivity Ablation Study Results.} We compare the SFT-only variant and our full Stage III model (\ourmodel) under standard inference and when forced to generate from an injected incorrect reasoning prefix. The full model exhibits substantially less performance degradation, demonstrating that Stage III training internalizes visually grounded constraints to resist incorrect intermediate textual logic.}
    \label{tab:exp_prefix}

    \setlength{\tabcolsep}{6pt}
    \renewcommand\arraystretch{1.15}
    \rowcolors{2}{gray!10}{white}
    \resizebox{\linewidth}{!}{%
      \begin{tabular}{lcccccc}
        \toprule
        \rowcolor{gray!25}
        \textbf{Variant} & \textbf{Fluid Intel.} & \textbf{Crystallized Intel.} & \textbf{Visuospatial Cog.} & \textbf{Mental Simu.} & \textbf{Visual Rout.} & \textbf{Avg.} \\
        \midrule
        Qwen3-VL-8B SFT \textit{w/} LVIP forced incorrect inference prefix & 28.6 & 45.4 & 26.5 & 31.3 & 25.8 & 34.7 \\
        Qwen3-VL-8B SFT \textit{w/} LVIP (standard inference) & 55.4 & 88.6 & 61.1 & 61.3 & 44.1 & 68.1 \\
        \midrule
        \ourmodel forced incorrect inference prefix & 51.1 & 78.5 & 54.9 & 56.0 & 40.9 & 61.4 \\
        \ourmodel (standard inference) & 63.8 & 91.0 & 69.0 & 68.0 & 50.5 & 73.8 \\
        \bottomrule
      \end{tabular}
    }
\end{table*}

\paragraph{Ablation Study on Prefix Sensitivity} To directly test whether LVIP is merely passively following the textual chain, we perform a robustness experiment in which we inject an incorrect intermediate inference prefix and force the model to continue generation from that point. As shown in~\Cref{tab:exp_prefix}, if LVIP only acted as a passive auxiliary signal, then the model would largely inherit the erroneous reasoning and collapse similarly to the SFT-only variant. Instead, we observe a much smaller degradation for the full Stage III model compared to Qwen3-VL-8B SFT \textit{w/} LVIP. This indicates that Stage III training with LVIP internalizes visually grounded constraints into the learned policy, making the model substantially more robust to flawed intermediate reasoning prefixes.

\paragraph{Ablation Study on Intermediate State Perturbation} If a model relies on an answer-conditioned shortcut, its intermediate steps should not causally drive the reasoning process. Therefore, corrupting an intermediate state should not disrupt a shortcut that directly maps to the final sequence. Following~\citep{li2026dynamics}, we test whether each latent visual state is functionally necessary by selecting successful reasoning trajectories and injecting random noise into the latent visual state at a specific step (Step 30). The model then continues decoding from the perturbed state.
We measure the MSE between the intermediate latent visual imagery $\hat{\mathbf{h}}_{y,i}$ at different CoT steps and the final ground-truth latent visual imagery $\mathbf{h}_y$, averaged over 10 sampled trajectories. As shown in~\Cref{tab:exp_noise}, the trajectory remains stable before the intervention. Once the Step-30 state is perturbed, downstream latent error increases, and the trajectory fails to reconverge to the clean path. This provides evidence that intermediate latent visual states play a causal role in sustaining the reasoning trajectory, rather than merely being correlated with answer confidence.

\paragraph{Ablation Study on Latent State Intervention}
To further test whether LVIP actively steers reasoning, we conduct a latent patching experiment inspired by~\citep{liang2026latent}. We intervene during intermediate decoding by swapping the latent visual state between correct and incorrect reasoning runs. We measure the change in the probability of generating the correct answer after the intervention ($\Delta p = p_\mathrm{after} - p_\mathrm{before}$). As shown in~\Cref{tab:exp_patching}, when we patch a distractor's latent state into a correct trajectory, the downstream rationale shifts toward that distractor. Conversely, patching a correct latent state into an incorrect run actively shifts the trajectory toward the correct answer. This provides strong evidence that the patched intermediate state carries vital causal information and actively steers the inference.

\paragraph{Ablation Study on Progressive Information Emergence} We also measure how predictive the intermediate latent visual state is of the final answer at different reasoning steps using a linear probe~\citep{alain2016understanding}. Specifically, we freeze \ourmodel and train a lightweight linear classifier to predict the final correct option exclusively using the intermediate latent visual imagery $\hat{\mathbf{h}}_{y, i}$ extracted at different steps of the reasoning process. If the model were taking a trivial shortcut, the latent state would immediately identify the answer from the start. As shown in~\Cref{tab:exp_probing}, the accuracy increases gradually and progressively rather than appearing fully formed at early steps. This rules out the trivial shortcut hypothesis, providing evidence that the target information is genuinely constructed step-by-step via latent visual imagery as the textual logic unfolds.

Altogether, these experiments provide evidence that LVIP learns meaningful intermediate visual states, rather than functioning as an answer-conditioned shortcut or a passive byproduct of textual CoT. \looseness=-1

\section{Ablation Study on the Number of Sampled Latent Rationales}
\label{appx_sec:ablation}
\setcounter{figure}{0}
\setcounter{table}{0}

\begin{table*}[!t]
    \centering
    \caption{\textbf{Rationales Quantity Ablation Study Results.} We report the average accuracy and average inference latency across different values of $n$. The default setting ($n=8$) provides an optimal trade-off, securing performance gains before latency increases non-linearly.}
    \label{tab:exp_n}
    \setlength{\tabcolsep}{6pt}
    \renewcommand\arraystretch{1.15}
    \rowcolors{2}{gray!10}{white}
    \resizebox{0.60\linewidth}{!}{%
      \begin{tabular}{lcc}
        \toprule
        \rowcolor{gray!25}
        \textbf{$n$} & \textbf{Average Accuracy} & \textbf{Average Latency} \\
        \midrule
        1 (No aggregation) & 71.4 & 1.5s \\
        4 & 72.9 & 1.8s \\
        8 (Default) & 73.8 & 2.4s \\
        16 & 73.9 & 3.8s \\
        32 & 73.9 & 6.5s \\
        \bottomrule
      \end{tabular}
    }
\end{table*}

To analyze the sensitivity of the hyperparameter $n$ (the number of sampled latent rationales) in the paragraph \textbf{Bayesian Posterior over Latent Rationales} in~\Cref{sec:stage_3} and its impact on inference latency, we conduct an ablation study by varying $n$ during inference. As shown in~\Cref{tab:exp_n}, increasing the number of sampled latent rationales gradually increases the accuracy. After $n=8$, the improvement in accuracy begins to level off and comes with a non-linear increase in inference latency. The default setting $n=8$ provides a robust balance between predictive performance and practical deployment efficiency.

\section{Ablation Study on Rationale Sampling Strategies}
\label{appx_sec:strategy}
\setcounter{figure}{0}
\setcounter{table}{0}

To ensure that the performance gains of \ourmodel do not simply come from a larger sampling budget of rationales during inference, and to explore the effectiveness of different sampling strategies, we conduct a comprehensive ablation study comparing various sampling and selection strategies. To ensure a fair comparison, we fix the total sampling budget for all multi-sample baselines to $n=8$. We maintain consistency across all experiments by using the same decoding temperature, maximum rationale length, and the same frozen scorer for reranking when applicable.

We evaluate the following strategies to isolate the impact of our proposed approach:

\begin{itemize}
    \item \textbf{Greedy Decoding}: A single deterministic decoding baseline representing the standard performance without multiple samples.
    \item \textbf{SC-Majority}: Following the Self-Consistency method from~\citep{wang2022self}, we sample $n=8$ rationales and perform a majority vote on the final answers, treating each path with equal weight. \looseness=-1
    \item \textbf{SC-Reranking}: We sample $n=8$ rationales and use the frozen scorer to select the path with the highest reward score, using only that path's final answer. \looseness=-1
    \item \textbf{Ours}: Our proposed sampling strategy described in~\Cref{sec:stage_3}.
\end{itemize}

\begin{table*}[!t]
    \centering
    \caption{\textbf{Sample Strategies Ablation Study Results.} All $n\!=\!8$ variants use the same number of sampled rationales. \ourmodel remains strongest, showing that the gain is not merely due to self-consistency or reranking.\looseness-1}

    \label{tab:strategy}
    \setlength{\tabcolsep}{6pt}
    \renewcommand\arraystretch{1.15}
    \rowcolors{2}{gray!10}{white}
    \resizebox{0.99\linewidth}{!}{%
        \begin{tabular}{llcccccc}
            \toprule
            \rowcolor{gray!25}
            \textbf{Variant} & \textbf{Strategy} & \textbf{Fluid Intel.} & \textbf{Crystallized Intel.} & \textbf{Visuospatial Cog.} & \textbf{Mental Simu.} & \textbf{Visual Rout.} & \textbf{Avg.} \\
            \midrule
            Qwen3-VL-8B (Base) & Greedy Decoding & 31.2 & 34.8 & 31.0 & 45.3 & 40.9 & 35.5 \\
            Qwen3-VL-8B (Base) & SC-Majority & 31.9 & 35.3 & 31.9 & 46.0 & 41.9 & 36.2 \\
            Qwen3-VL-8B (Base) & SC-Reranking & 32.2 & 35.9 & 31.9 & 46.7 & 41.9 & 36.6 \\
            \midrule
            Qwen3-VL-8B SFT \textit{w/o} LVIP & Greedy Decoding & 51.1 & 76.6 & 63.7 & 59.3 & 41.9 & 62.3 \\
            Qwen3-VL-8B SFT \textit{w/o} LVIP & SC-Majority & 52.2 & 77.7 & 64.6 & 60.0 & 41.9 & 63.2 \\
            Qwen3-VL-8B SFT \textit{w/o} LVIP & SC-Reranking & 52.9 & 78.3 & 64.6 & 60.0 & 43.0 & 63.7 \\
            \midrule
            Qwen3-VL-8B SFT \textit{w/} LVIP & Greedy Decoding & 55.4 & 88.6 & 61.1 & 61.3 & 44.1 & 68.1 \\
            Qwen3-VL-8B SFT \textit{w/} LVIP & SC-Majority & 56.2 & 88.9 & 61.9 & 62.0 & 45.2 & 68.7 \\
            Qwen3-VL-8B SFT \textit{w/} LVIP & SC-Reranking & 56.5 & 89.1 & 62.8 & 62.0 & 45.2 & 69.0 \\
            \midrule
            Qwen3-VL-8B SFT \textit{w/} LVIP + GRPO & Greedy Decoding & 59.1 & 89.9 & 64.6 & 65.3 & 46.2 & 70.8 \\
            Qwen3-VL-8B SFT \textit{w/} LVIP + GRPO & SC-Majority & 59.4 & 90.2 & 65.5 & 66.0 & 47.3 & 71.3 \\
            Qwen3-VL-8B SFT \textit{w/} LVIP + GRPO & SC-Reranking & 59.8 & 90.2 & 65.5 & 66.0 & 47.3 & 71.4 \\
            \midrule
            Qwen3-VL-8B SFT \textit{w/} LVIP + GRPO + LR & Greedy Decoding & 59.8 & 90.2 & 65.5 & 66.7 & 47.3 & 71.5 \\
            Qwen3-VL-8B SFT \textit{w/} LVIP + GRPO + LR & SC-Majority & 60.5 & 90.5 & 66.4 & 67.3 & 48.4 & 72.1 \\
            Qwen3-VL-8B SFT \textit{w/} LVIP + GRPO + LR & SC-Reranking & 61.2 & 90.8 & 66.4 & 67.3 & 48.4 & 72.4 \\
            \midrule
            \rowcolor{gray!15} \ourmodel (Ours) & Ours & \textbf{63.8} & \textbf{91.0} & \textbf{69.0} & \textbf{68.0} & \textbf{50.5} & \textbf{73.8} \\
            \bottomrule
        \end{tabular}
    }
\end{table*}

As shown in~\Cref{tab:strategy}, across all variants from the base model up to GRPO models, sampling strategies such as Self-Consistency and Scorer Reranking consistently provide marginal improvements over Greedy Decoding. However, these improvements show diminishing returns. For instance, applying SC-Reranking to the standard SFT models cannot close the performance gap to models trained with trajectory-level optimization (GRPO). Furthermore, even when applying the SC-Reranking strategy to our strongest baseline (SFT \textit{w/} LVIP + GRPO + LR), the average performance (72.4\%) still falls short of \ourmodel (73.8\%). This firmly establishes that the superior performance of \ourmodel is driven intrinsically by our visually grounded latent rationale optimization and targeted sampling framework, rather than merely relying on a larger sampling budget.

\section{Ablation Study on Teacher Models}
\label{appx_sec:teacher}
\setcounter{figure}{0}
\setcounter{table}{0}

We regenerate the full set of training rationales using Gemini 2.5 Pro~\citep{comanici2025gemini25pushingfrontier} instead of GPT-5~\citep{singh2025openai} while keeping all other experimental settings unchanged. This yields an average accuracy of 73.5, compared to the 73.8 achieved with GPT-5-generated rationales. The near-identical performance indicates that our results are robust and not specific to the rationale style of a single teacher model, demonstrating that our method learns generalizable visual cognition patterns rather than merely adapting to distributions of specific datasets or teacher-generated rationales.

\section{Ablation Study on Backbone Models}
\label{appx_sec:backbone}
\setcounter{figure}{0}
\setcounter{table}{0}

To evaluate the generalizability of our proposed Cognitive Supersensing approach, we extend our experiments to incorporate more advanced vision-language foundation models. Specifically, we replace the original backbone with InternVL3.5-8B~\citep{wang2025internvl3} and Qwen3.5-9B~\citep{qwen35blog}. We retrain these new baseline models and comprehensively re-evaluate them on CogSense-Bench to isolate the impact of our method across different model configurations.

The results, detailed in~\Cref{tab:exp_backbones}, demonstrate that Cognitive Supersensing consistently and significantly improves performance across all evaluated cognitive dimensions, regardless of the underlying backbone model. Notably, applying our method to Qwen3.5-9B yields a remarkable average score of \textbf{78.3}, representing an absolute improvement of \textbf{9.2} points over the standard fine-tuned Qwen3.5-9B baseline (69.1). Similarly, CogSense augments the InternVL3.5-8B model by \textbf{11.3} points on average compared to its fine-tuned counterpart. These findings confirm that our approach is not constrained by specific architectural choices, but rather serves as a broadly effective mechanism for enhancing latent visual reasoning and multifaceted cognitive capabilities in state-of-the-art MLLMs.

\begin{table*}[!t]
  \centering
    \caption{\textbf{Backbone Ablation Study Results.} We evaluate the generalizability of Cognitive Supersensing across different foundation models. The results demonstrate that Cognitive Supersensing consistently yields significant performance gains regardless of the backbone architecture. Red numbers (\red{\cdot}) denote the absolute improvement over the corresponding fine-tuned (FT) baseline.}

  \label{tab:exp_backbones}

  \setlength{\tabcolsep}{5.3pt}
  \renewcommand{\arraystretch}{1.12}
  \footnotesize

  \resizebox{0.98\linewidth}{!}{%
  \begin{tabular}{lcccccc}
    \toprule
    \rowcolor{gray!25}
    \textbf{Model} & \textbf{Fluid Intel.} & \textbf{Crystallized Intel.} & \textbf{Visuospatial Cog.} & \textbf{Mental Simu.} & \textbf{Visual Rout.} & \textbf{Avg.} \\
    \midrule

    Qwen3-VL-8B (FT) & 51.1 & 76.6 & 63.7 & 59.3 & 41.9 & 62.3 \\
    \rowcolor{gray!8}
    InternVL3.5-8B (FT) & 52.2 & 80.2 & 63.7 & 59.3 & 41.9 & 63.9 \\
    Qwen3.5-9B (FT) & 56.5 & 89.1 & 61.9 & 62.7 & 46.2 & 69.1 \\

    \midrule
    CogSense (Qwen3-VL-8B) &~~~~~~~~63.8\red{12.7} &~~~~~~~~91.0\red{14.4} &~~~~~~{69.0}\red{5.3} &~~~~~~68.0\red{8.7} &~~~~~~50.5\red{8.6} &~~~~~~~73.8\red{11.5} \\
    \rowcolor{gray!8}
    CogSense (InternVL3.5-8B) &~~~~~~~~{64.1}\red{11.9} &~~~~~~~~{93.2}\red{13.0} &~~~~~~{70.8}\red{7.1} &~~~~~~~~{69.3}\red{10.0} &~~~~~~{51.6}\red{9.7} &~~~~~~~{75.2}\red{11.3} \\
    CogSense (Qwen3.5-9B) &~~~~~~~~{69.9}\red{13.4} &~~~~~~~{97.3}\red{8.2} &~~~~~~~68.1\red{6.2} &~~~~~~{70.0}\red{7.3} &~~~~~~{53.8}\red{7.6} &~~~~~~{78.3}\red{9.2} \\
    \bottomrule
  \end{tabular}%
  }
\end{table*}

\section{OOD Evaluation on Open-Ended VQA Benchmarks}
\label{appx_sec:ood}
\setcounter{figure}{0}
\setcounter{table}{0}

\begin{table}[!t]
    \centering
    \caption{\textbf{Out-of-Domain Evaluation on Open-Ended VQA Benchmarks}. We evaluate the generalization ability on open-ended VQA tasks of \ourmodel compared with the base model.}
    \label{tab:exp_ood}

    \setlength{\tabcolsep}{4pt}
    \renewcommand\arraystretch{1.15}
    \rowcolors{2}{gray!10}{white}
    \resizebox{\linewidth}{!}{%
      \begin{tabular}{l ccccc cc}
        \toprule
        \rowcolor{gray!25}
        \textbf{Model} & \multicolumn{5}{c}{\textbf{BabyVision}} & \multicolumn{1}{c}{\textbf{MathVerse}} & \multicolumn{1}{c}{\textbf{CLEVRER}} \\
        \rowcolor{gray!25}
         & Fine-grained Discrimination & Visual Tracking & Spatial Perception & Visual Pattern Recognition & All & w/o & Descriptive \\
        \midrule
        Qwen3-VL-8B (Base) & 12.7 & 10.8 & 15.0 & 15.0 & 13.1 & 62.1 & 62.4 \\
        \ourmodel (Ours) & 15.3 & 13.2 & 18.1 & 17.6 & 15.7 & 65.7 & 67.9 \\
        \bottomrule
      \end{tabular}
    }
\end{table}

The LVIP training objective uses candidate answer images to supervise the latent visual target and learn how to generate latent visual features. However, at inference time, LVIP does not require access to candidate answer images or any explicit multiple-choice options. Instead, it internally generates latent visual imagery that supports the model's textual reasoning, enabling the same mechanism to be applied to both multiple-choice and open-ended tasks.
To verify the open-ended generalization of \ourmodel, we conduct an OOD evaluation on open-ended VQA benchmarks, including BabyVision~\citep{chen2026babyvision}, MathVerse~\citep{zhang2024mathverse}, and CLEVRER~\citep{yi2019clevrer}-Descriptive. Reported values represent the Pass@1 accuracy, following the settings of the original benchmarks. These evaluations contain no candidate images to select or match at inference. Thus, their gains cannot be explained by an inference-time mechanism that simply aligns the hidden state with option images. As shown in~\Cref{tab:exp_ood}, our model consistently outperforms the baseline even on strictly open-ended VQA tasks, demonstrating that the cognitive reasoning abilities gained through image-based supervision transfer effectively to open-ended settings.

\section{Broader Impacts}
\label{appx_sec:impact}

This paper presents work whose goal is to advance the fields of machine learning and cognitive science, specifically in multimodal reasoning and visual cognition. The proposed methods aim to improve MLLMs' ability to solve complex visual reasoning tasks, which could have broad applications in education, scientific discovery, and assistive technologies. While we believe the potential benefits are substantial, we acknowledge that improved visual reasoning capabilities could be misused in contexts such as automated surveillance or manipulation. We encourage the research community to consider ethical implications and develop appropriate safeguards as these technologies advance.

\section{Limitations}
\label{appx_sec:limitations}
While \ourmodel achieves strong visual cognitive performance at the 8B scale, our experiments focus on validating the proposed Cognitive Supersensing paradigm within a moderate-scale MLLM setting. Future work could investigate scaling properties to substantially larger 30B- or 70B-style MLLM backbones, larger training corpora, and distributed training setups.

\end{document}